\documentclass[preprint,12pt,sort&compress]{elsarticle}

\usepackage{amssymb}
\usepackage{subfigure}
\usepackage{subfigure}
\usepackage{graphicx}
\usepackage{color}
\usepackage[graphicx]{realboxes}
\usepackage{makecell}

\usepackage[colorlinks, linkcolor=blue, anchorcolor=blue, citecolor=blue]{hyperref}
\usepackage{bm}
\usepackage{booktabs}
\usepackage{amsmath}
\usepackage{lipsum}
\usepackage{tabularx, booktabs,multirow}
\renewcommand{\appendixname}{APPENDIX}
\usepackage[colorlinks,linkcolor=blue]{hyperref}

\journal{Journal of Computational Physics}

\begin{document}

\begin{frontmatter}



\title{A hybrid  Decoder-DeepONet operator regression framework for unaligned observation data}


\author[inst1]{Bo Chen}
\author[inst1]{Chenyu Wang }
\author[inst1]{Weipeng Li\corref{cor1}}
\author[inst2]{Haiyang Fu}

\affiliation[inst1]{organization={School of Aeronautics and Astronautics},
            addressline={Shanghai Jiao Tong University}, 
            city={Shanghai},
            postcode={200240}, 
            state={},
            country={China}}
        
\affiliation[inst2]{organization={School of Information Science and Engineering},
	addressline={Fudan University}, 
	city={Shanghai},
	postcode={200433}, 
	state={},
	country={China}}

\cortext[cor1]{Corresponding author, E-mail address: liweipeng@sjtu.edu.cn (Weipeng Li).}

\begin{abstract}
Deep neural operators (DNOs) have been  utilized to approximate nonlinear mappings between function spaces. 
However, DNOs face the challenge of increased dimensionality and computational cost associated with unaligned observation data. 
In this study, we propose a hybrid Decoder-DeepONet operator regression framework to handle unaligned data effectively. 
Additionally, we introduce a Multi-Decoder-DeepONet, which utilizes an average field of the training data as input augmentation. 
The consistencies of the frameworks with the operator approximation theory are provided,  on the basis of the universal approximation theorem.
Two numerical experiments, Darcy problem and flow-field around an airfoil, are conducted to validate the efficiency and accuracy  of the proposed methods. Results illustrate the advantages of Decoder-DeepONet and Multi-Decoder-DeepONet in handling unaligned observation data and showcase their potentials in improving prediction accuracy.

\end{abstract}


\begin{highlights}
\item 
Efficient Operator Regressions: Decoder-DeepONet and Multi-Decoder-DeepONet frameworks outperform traditional DNOs, enhancing accuracy and efficiency in handling unaligned data.
\item 
Input Augmentation Advancement: Multi-Decoder-DeepONet integrates averaged flow fields, boosting prediction accuracy in complex nonlinear mappings.
\item 
Consistent Operator Approximation: Hybrid frameworks maintain alignment with theory, combining decoder nets for improved handling of unaligned data problems.
\end{highlights}

\begin{keyword}
deep neural operators \sep universal approximation theorem \sep unaligned data \sep Decoder-DeepONet
\PACS 0000 \sep 1111
\MSC 0000 \sep 1111
\end{keyword}

\end{frontmatter}



\section{Introduction}

Computer-assisted artificial intelligence (AI) has become a powerful tool for solving a wide range of forward and inverse problems\cite{Ricardo,bruntonMachineLearningFluid2019,kutzDeepLearningFluid2017,RAISSIPINN,chenPhysicsinformedMachineLearning2021,hasegawaCNNLSTMBasedReduced2020,hijaziDatadrivenPODGalerkinReduced2020,reichsteinDeepLearningProcess2019,larioNeuralnetworkLearningSPOD2022}.
In recent years, building surrogate models to solve partial differential equations (PDEs) with deep neural networks (DNNs)\citep{long2018pdenet,longPDENetLearningPDEs2019,ExtractingStructuredDynamicalSystems,ChangShu}, such as convolutional neural networks (CNNs), recurrent neural networks (RNNs), long short-term memory networks (LSTMs), generative adversarial networks (GANs), and physics-informed neural networks (PINNs), has been achieved and tested to predict flow fields\cite{hijaziDatadrivenPODGalerkinReduced2020}, compute chemical reactions\cite{chemistry}, and solve protein folding problems\cite{noeMachineLearningProtein2020}, to name a few. 
The surrogate models may significantly reduce the computational cost once they are well-trained\cite{bruntonMachineLearningFluid2019}. Most of the DNNs rely on the universal approximation theorem\cite{cybenkotApproximationSuperpositionsSigmoidal}, which may approximate arbitrary functions with arbitrary accuracy.
However, these models have limitations in accurately predicting the behavior of new systems subjected to different input signals, including boundary conditions, initial conditions, and source terms\cite{luLearningNonlinearOperators2021}. 

 In contrast to DNNs that approximate functions, deep neural operators (DNO) are capable of approximating nonlinear mappings from an infinite function space to another, with the universal approximation theorem of operators providing the necessary mathematical guarantee \cite{chen1995}.
 The first framework of DNO was proposed by \citet{luLearningNonlinearOperators2021} in 2019, and named deep operator net (DeepONet). It  consists of a branch net receiving function inputs and a trunk net receiving temporal and spatial observations. 
  In 2020, \citet{li_fourier_2021} formulated a new neural operator, Fourier neural operator (FNO), in which the integral kernel is directly parameterized  in Fourier space.
  Both DeepONet and FNO have been demonstrated to learn operator mappings and  forecast the evolution of systems with the variation of input signals\cite{luComprehensiveFairComparison2022}.  
  DeepONet is more widely extended and utilized due to its flexible structure and small generalization errors. 
 Using proper orthogonal decomposition (POD) to  extract energetic modes in the trunk net,  \citet{luComprehensiveFairComparison2022} further proposed a  POD-DeepONet, in which the branch net is just utilized to learn the coefficients of the POD modes, resulting in less learning pressure and increased prediction accuracy.
 To extend the input function space from single to multiple, a multiple-input operator network (MIOnet)\cite{jinMIONetLearningMultipleInput2022a} consisting of more than one branch net was developed.
  Another impressive extension is physics-informed DeepONet\cite{RAISSIPINN}, which provides a semi-supervised type of learning through a combination of PINNs and DeepONet. 
  Further extensions of DeepOnet include multiple pre-trained DeepONets\cite{CAI2021110296}, DeepONet with uncertainty quantification\cite{lin2021accelerated}, multiscale DeepONet\cite{liu2021multiscale}, and others\cite{Yang_2022,liu2022deeppropnet}. 
  DeepONet and its varieties have been applied in a broad range of applications, including the prediction of disturbance evolution in hypersonic boundary layers\cite{dileoni2021deeponet}, the investigation of coupled flow and finite-rate chemistry in shock flow fields\cite{maoDeepMMnetHypersonics2021}, the prediction of progressive intramural damage culminating in aortic dissection\cite{Yin2021}, and multiscale modeling of mechanical problems\cite{Yin_2022}.

\begin{figure}
\centering 
\subfigure[\scriptsize $k_{1}(\xi)$]
{\includegraphics[width=0.45\textwidth,trim=20 190 210 20,clip]{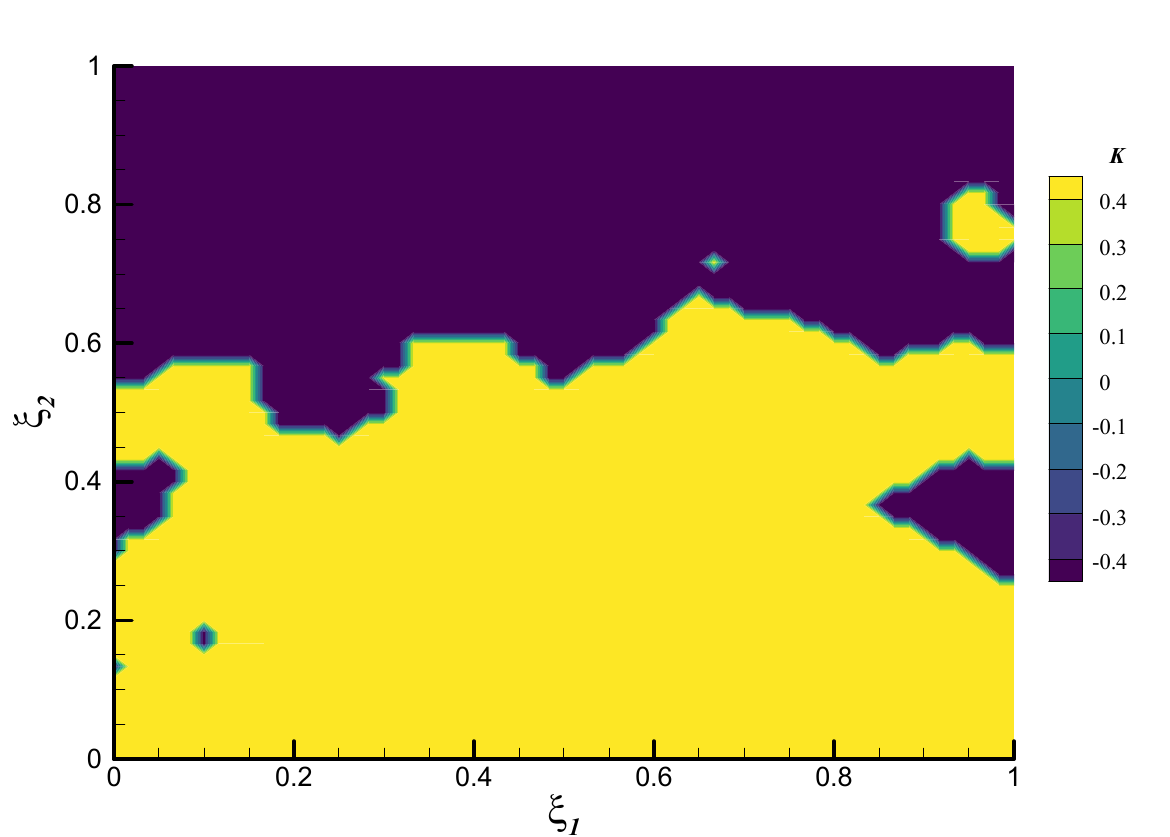}}
\subfigure[\scriptsize$k_{2}(\xi)$]
{\includegraphics[width=0.45\textwidth,trim=20 190 210 20,clip]{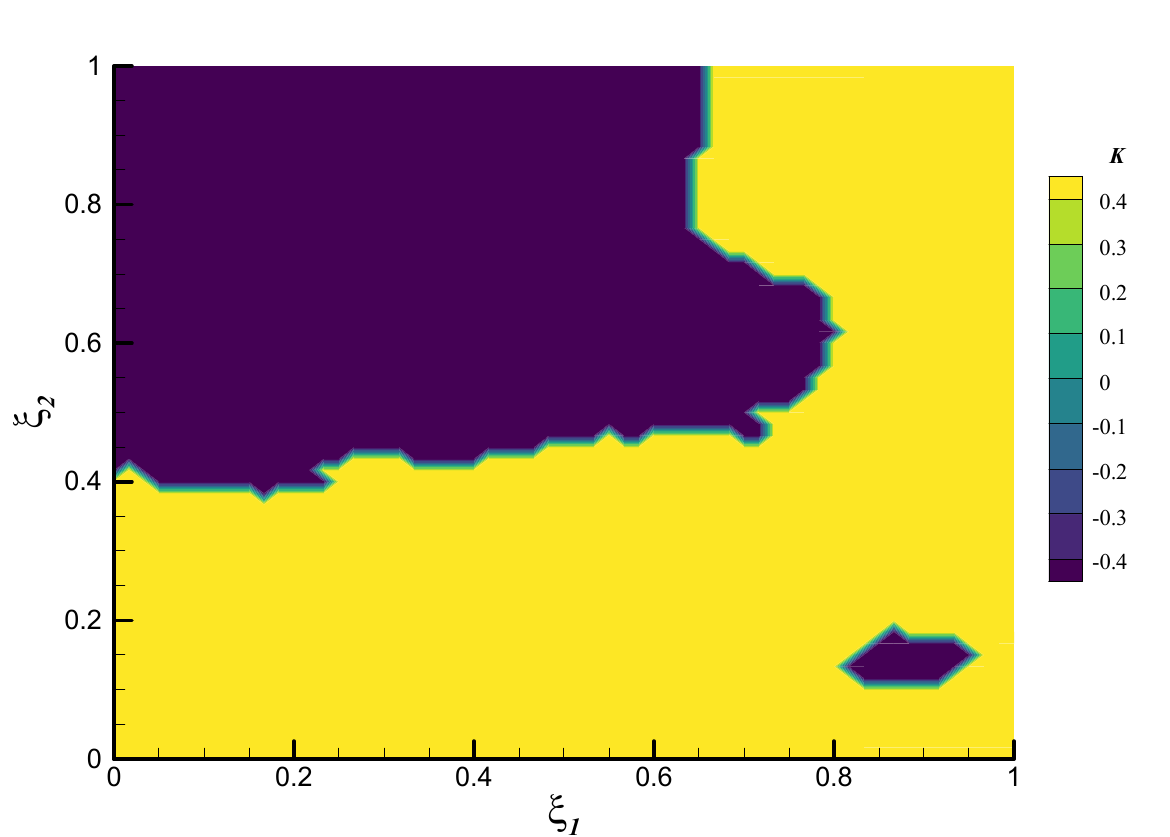}}
\subfigure[\scriptsize $h_{1}(k_{1})(\xi)$-aligned]
{\includegraphics[width=0.45\textwidth,trim=20 190 210 20,clip]{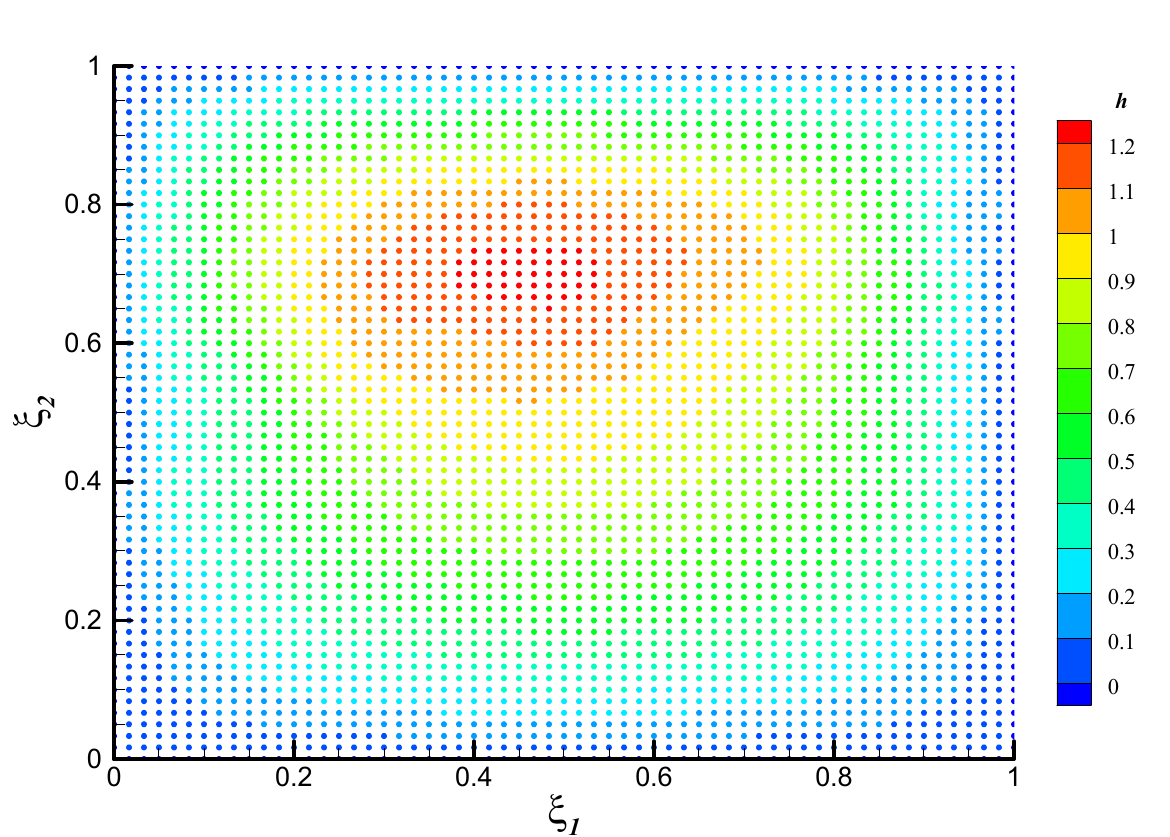}}
\subfigure[\scriptsize$h_{2}(k_{2})(\xi)$-aligned]
{\includegraphics[width=0.45\textwidth,trim=20 190 210 20,clip]{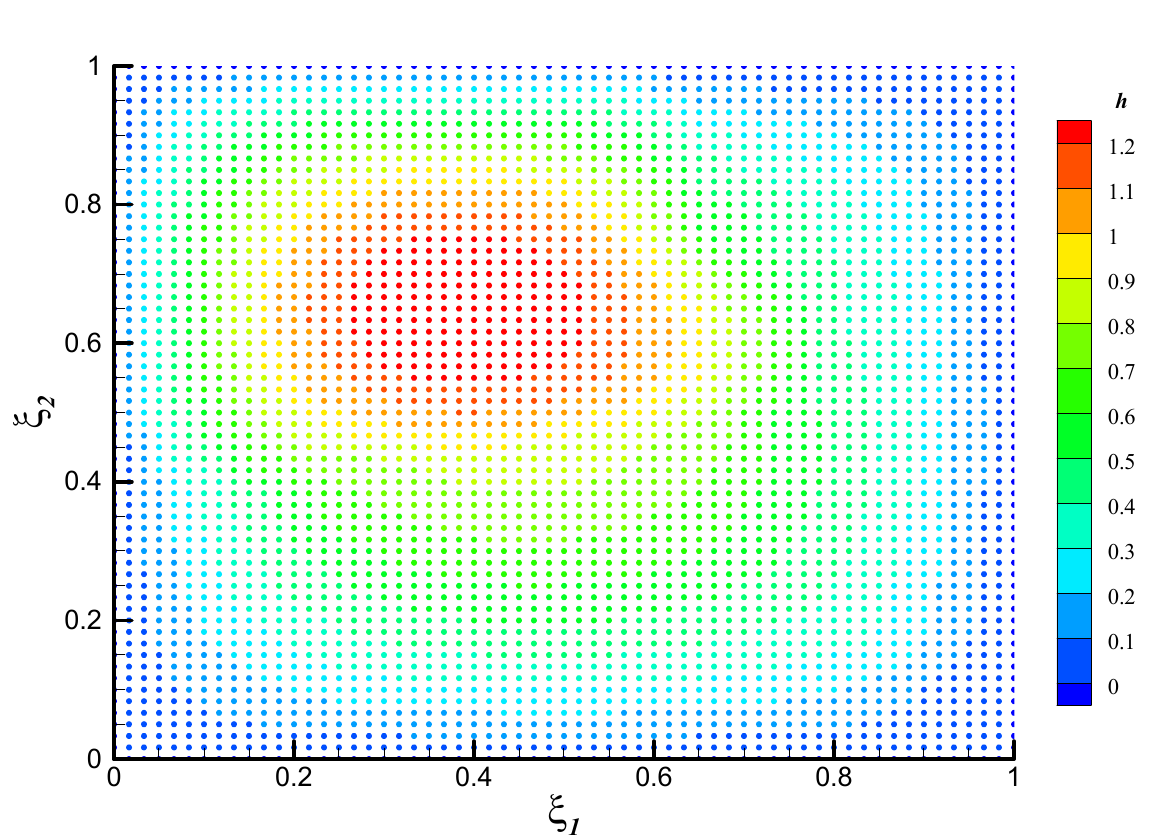}}
\subfigure[\scriptsize $h_{1}(k_{1})(\xi)$-unaligned]
{\includegraphics[width=0.45\textwidth,trim=20 190 210 20,clip]{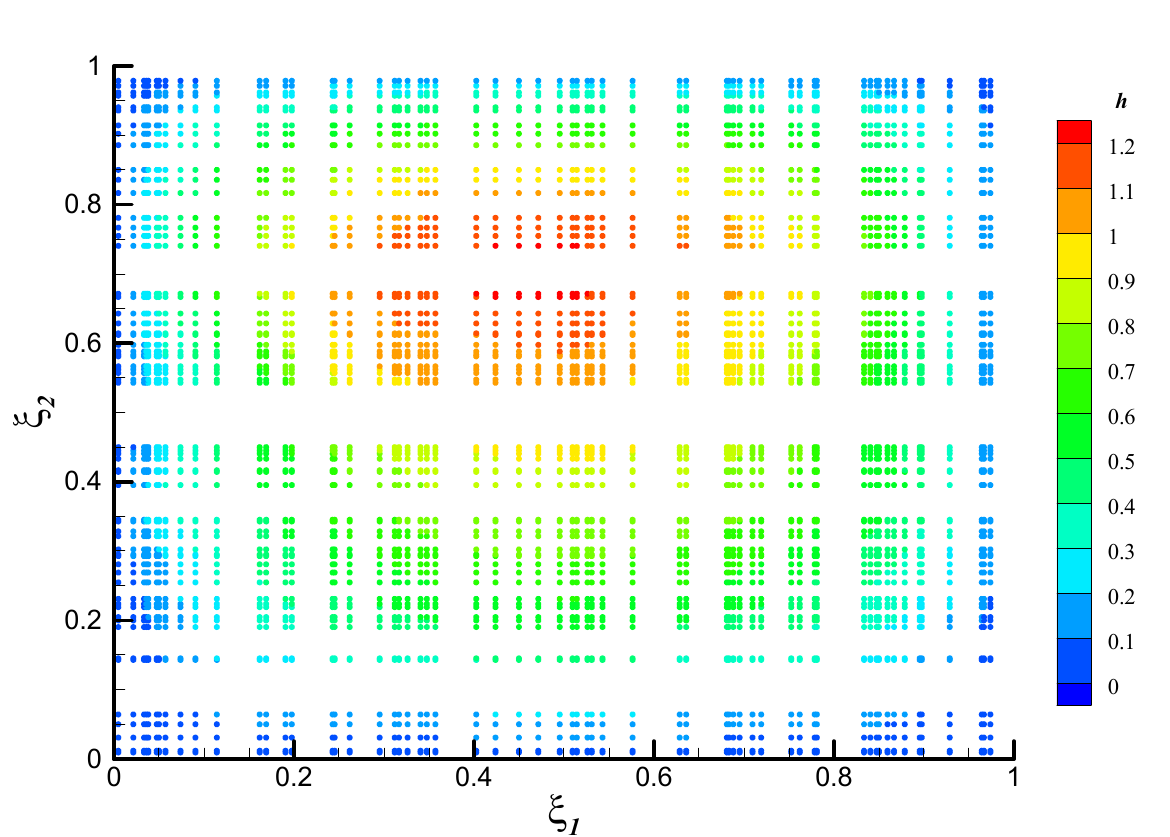}}
\subfigure[\scriptsize$h_{2}(k_{2})(\xi)$-unaligned]
{\includegraphics[width=0.45\textwidth,trim=20 190 210 20,clip]{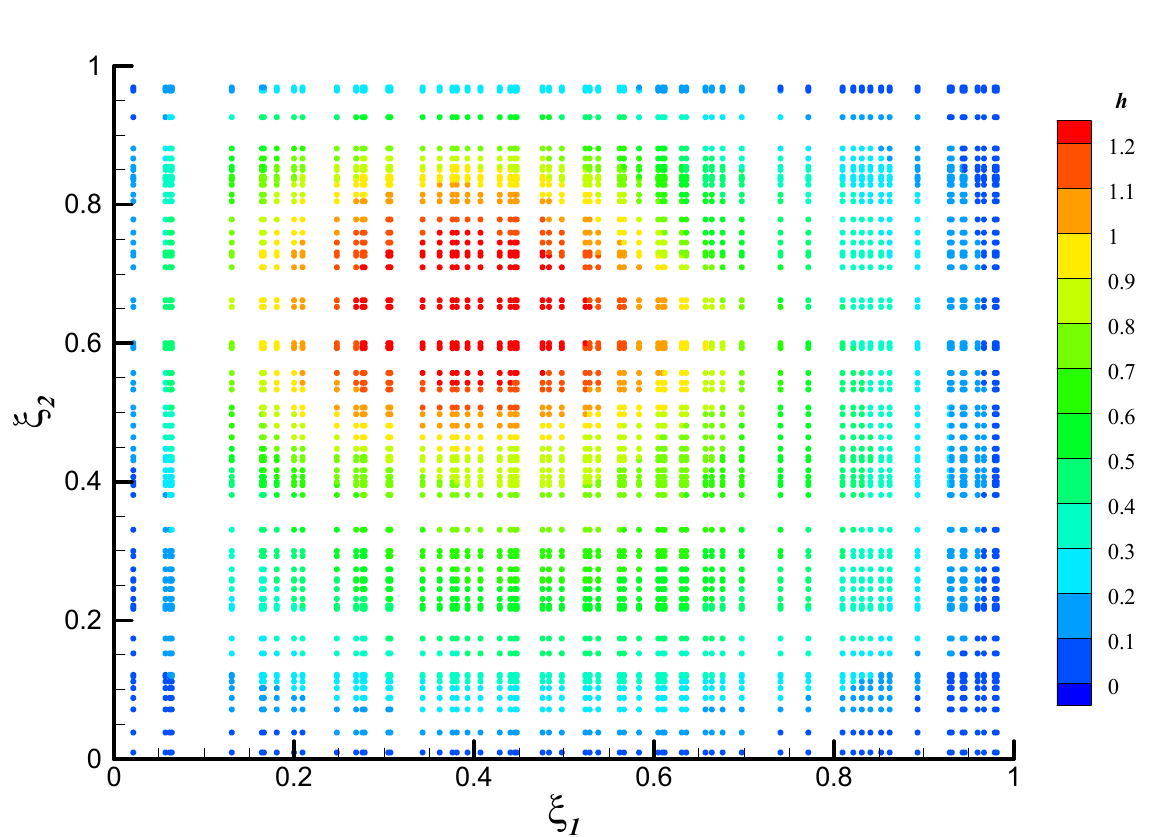}}
\caption{The comparison of aligned and unaligned dataset. (a) A permeability field $k_{1}(\xi)$. (b) Another permeability field $k_{2}(\xi)$. (c) The pressure fields $h_{1}(k_{1})(\xi)$ corresponding to $k_{1}(\xi)$ with a uniform distribution of observations . (d) The pressure fields $h_{2}(k_{2})(y)$ corresponding to $k_{2}(\xi)$ with a uniform distribution. (e) $h_{1}(k_{1})(\xi)$ with a random distribution. (f) $h_{2}(k_{2})(\xi)$ with a random distribution. }
\label{fig:Darcy}
\end{figure}

Two types of observation data, namely aligned  and unaligned data, are usually encountered in academic studies and practical applications. 
The former represents that the physical coordinates of observations (outputs) are fixed under different input conditions. The unaligned data is formed when the distribution of the observations varies with the input conditions. 
For instance, the two-dimensional (2D) Darcy problem involves the mapping from a permeability field $k(\xi)$ to a pressure field $h(k)(\xi)$, where $\xi$ is the 2D observations $(\xi^{1},\xi^{2})$. This problem can be described using either aligned or unaligned data sets. Figure \ref{fig:Darcy} illustrates the differences by using aligned  and unaligned datasets. 
Figures \ref{fig:Darcy}($a, b$)  display two distributions of the permeability fields $k_{1}$ and $k_{2}$, which are input conditions. Correspondingly, with the pressure fields observed with aligned dataset are shown in figures \ref{fig:Darcy}($c, d$), and those observed with unaligned dataset are shown in figures \ref{fig:Darcy}($e,f$). 
It can be seen that under different permeability fields as inputs, the observation locations of the pressure fields keep constant (uniformly distributed here) in figures \ref{fig:Darcy}($c, d$), but varies (randomly distributed) in figures \ref{fig:Darcy}($e,f$). 
Despite employing an identical number of observation points, the observed pressure fields ($h_{1}$ or $h_{2}$) differ due to disparities in the locations of observation points, which may significantly impact the learning of mappings from $k$ to $h$.
Facing unaligned data, some DNNs, such as CNNs, lack the input of observation, which results in the learning of solely image-to-image mappings between initial conditions and results. Inherently, these approaches treat unaligned data as aligned data, disregarding the influences of observation points on the system variations.

Previous studies on DeepONet focused on the aligned data \cite{luComprehensiveFairComparison2022,dileoni2021deeponet,maoDeepMMnetHypersonics2021,Yin2021,Yin_2022}.  
However, in specific application scenarios, the use of aligned data is not suitable for the development of surrogate models. For example, the aerodynamic shape optimization of airfoils, rotors, wings, and turbine blades  involves the unaligned data problem, as the computational grids (flow field observations) vary with the shape configurations.
Actually, the classical DeepONet\cite{luLearningNonlinearOperators2021} is able to handle both aligned and unaligned data through different data processing methodologies. 
To address the aligned data, CartesianProd implementation\cite{luComprehensiveFairComparison2022} is employed. Fundamentally, this implementation reuses the same set of trunk net input for all branch net inputs,  reducing the training dataset size and computational redundancy. 
However, in the case of unaligned data, only the original prod implementation in Ref. \cite{luLearningNonlinearOperators2021} can be utilized, resulting in  much larger dimensionality and more computational cost than those of aligned data.
For high-dimensional unaligned observations, the exponentially increased dimensionality will challenge the network training, deteriorate the prediction accuracy, and heighten the memory requirements. 

In this article, we propose a novel hybrid Decoder-DeepONet operator regression framework to handle the unaligned observation data. 
Additionally, a multiple input form named Multi-Decoder-DeepONet, inspired by MIOnet, is also effective, in which the average field of training data is fed into another subnetwork as input argumentation. We consider the prediction of Darcy problem and the prediction of flow-field around an airfoil to demonstrate the advantages of Decoder-DeepONet and Multi-Decoder-DeepONet compared with traditional networks. 
The paper is organized as follows. In Section \ref{method}, we provide a brief introdcution of the architecture and data evolution of DeepONet and its varieties and introduce the novel Decoder-DeepONet and Multi-Decoder-DeepONet operator regression framework. Comparative analysis of Decoder-DeepONet and Multi-Decoder-DeepONet against traditional DNOs is performed in Section \ref{results}. The conclusions are highlighted in Section \ref{conclusion}.

\section{Method} \label{method}
In this section, we give a concise introduction to the conventional approximation theory applied to both DeepONet and POD-DeepONet. Subsequently, we introduce the novel Decoder-DeepONet and Multi-Decoder-DeepONet operator regression and furnish the proof of the approximation theory.

\subsection{The universal operator approximation theorem} \label{theory}

The universal operator approximation theorem was first proposed by \citet{chen1995}. 
 Here we define an operator $G$ receiving the input function $u(x)$. At the observation points $\xi$ in the domain of $G(u)$, $G(u)(\xi)$ denotes the output field to be approximated. 

Name that \textit{$K_{1}$} is a compact set in a Banach space, \textit{$K_{2}$} is a compact set in $\mathbb{R}^{d}$, $V$ is a compact set in \textit{$C(K_{1})$} which means the Banach space of all continuous functions defined on {$K_{1}$} with norm $ \Vert f \Vert_{C(K_{1})} = MAX_{x\in{K_{1}}} \mid f(x) \mid$. 
Then, the operator $G$ to be learned can be expressed as:
 \begin{equation}
     {G} : V \ni{u} \mapsto G(u)(\xi) \in {C(K_{2})}.
 \end{equation}
Based on the study of \citet{chen1995}, the universal operator approximation theorem is given:

\noindent \textbf{Theorem 1}: For any small value $\epsilon > 0$, there are positive integers \textit{n}, \textit{p} and \textit{m},  $\sigma\in$ Tauber-Wiener (TW) functions, constants $c_{j}^{k}$, $\xi_{ij}^{k}$, $\theta_{i}^{k}$, $\zeta_{k}\in \mathbb{R}$, $w_{k}\in \mathbb{R}^{d}$, $x_{j}\in K_{1}$, $i=1,...,n$, $k=1,...,p$, and $j=1,...,m$, such that:
\begin{equation}
     \left| G(u)(\xi) - \sum_{k=1}^{p}\sum_{i=1}^{n}c_{j}^{k}\sigma\left(\sum_{j=1}^{m}\xi_{ij}^{k}u(x_{j})+\theta_{i}^{k}\right)\sigma(w_{k}\cdot{\xi}+\zeta_{k}) \right| < \frac{1}{2}\epsilon
\end{equation}
holds for all $u\in V$ and $\xi\in K_{2}$, where $G:V\to C(K_{2})$ is the nonlinear continuous operator.

The universal operator approximation theorem  suggests that a neural network containing at least one hidden layer is able to approximate any nonlinear continuous functional or operator.
An architecture of operator neural network can be built to learn operator mappings from data, and the  data-driven approach holds great potentials in a wide range of applications. 

\subsection{DeepONet}\label{deeponet}

\citet{luLearningNonlinearOperators2021} extended the universal operator approximation theorem to a more general from. Considering two continuous vector functions $\textbf{b}(b_{1},...,b_{p}):\mathbb{R}^{m} \to \mathbb{R}^{p}$ and $\textbf{t}(t_{1},...,t_{p}):\mathbb{R}^{d} \to \mathbb{R}^{p}$, if we make
$b_{k}(x)=\sum_{i=1}^{n}c_{j}^{k}\sigma\left(\sum_{j=1}^{m}\xi_{ij}^{k}x_{j}+\theta_{i}^{k}\right)$ 
and
$t_{k}(\xi)=\sigma(w_{k}\cdot{\xi}+\zeta_{k})$, then Theorem 1 can be generalized as:

\noindent \textbf{Theorem 2}: For any $\epsilon > 0$, there are $x_{j}\in K_{1}$, such that:
\begin{equation}
     \left| G(u)(\xi) - \langle \textbf{b}(u(x_{1}),u(x_{1}),...,u(x_{m})) , \textbf{t}(\xi)\rangle \right| < \frac{1}{2}\epsilon
     \label{equ:3}
\end{equation}
holds for all $u\in V$ and $\xi \in K_{2}$, 
and $ \langle \cdot , \cdot \rangle$ means the dot product. 
Essentially, Theorem 2  extends shallow networks in Theorem 1 to deep networks, which contain the activation function $\sigma\in$ TW functions to satisfy the classical universal approximation theorem of functional\cite{256500}.



\begin{figure}
	\centering
	\includegraphics[width=1\textwidth,trim=160 290 305 155,clip]{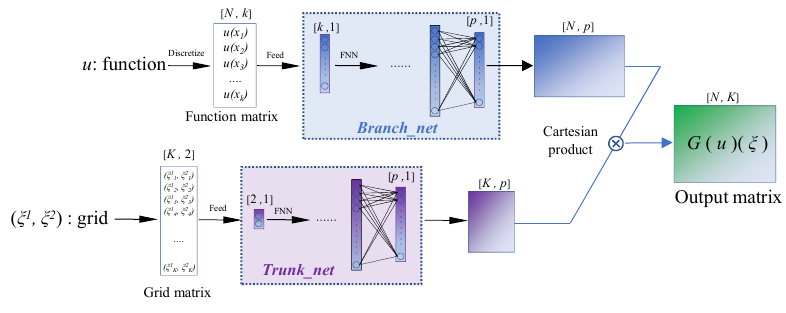}
	\caption{The architecture of DeepONet. Both branch net and trunk net are shown in the form of FNNs. The data fed into DeepONet is aligned here. $N$ is the number of input functions, each of which is dispersed by the $k$ points (u($x_1$),u($x_2$),u($x_3$) ....u($x_k$)); $K$ is the number of grid locations $\xi$ under each function $u$. To clearly display the datasize, $\xi$ in this paper is all shown as a 2D location $(\xi^{1},\xi^{2})$.}
	\label{fig:DeepOnet}
\end{figure}

Based on Theorem 2, a network architecture of DeepONet\cite{luLearningNonlinearOperators2021} was developed to realize the operator mapping. Figure \ref{fig:DeepOnet} shows the architecture of DeepONet.
It consists of a branch net and a trunk net. 
Note that FNNs are used in  the subnetwroks  as an example. 
The branch net receives functional inputs \textit{u}, encompassing initial conditions, boundary conditions, or source terms. The trunk net receives temporal or spatial observation coordinates $\xi$. With the dot product or Cartesian product (only for aligned data) operation, the output ${G} (u)(\xi)$ is approximated:
\begin{equation}
    {G} (\textit{u}) (\xi) \approx \sum_{k=1}^{p} branch_k (\textit{u}) \cdot trunk_k (\xi) + b_0 ,
    \label{bk}
\end{equation}
where $b_0$ is a bias to enhancing the representational capacity of the net. Note that the dimension evolution process for a two-dimensional aligned dataset is added in figure \ref{fig:DeepOnet}, making it different from the structure in Ref. \cite{luLearningNonlinearOperators2021}. 
For the aligned dataset, the function matrix and grid matrix have dimensions of $[N, k]$ and $[K, 2]$ to be fed into the branch and trunk net, respectively. A Cartesian product of the subnet output matrices is performed to get the forecasting output matrix $G(u)(\xi)$.
However, in the case of unaligned dataset, the dimension of function matrix and grid matrix will be $[N\times K,k]$ and $[N\times K,2]$, respectively. Instead of the Cartesian product, the forecasting output matrix $G(u)(\xi)$ will be obtained with a dot product operation of the two subnetwork output matrices, whose dimensions are $[N\times K,p]$. Obviously, the mapping relationship for the unaligned dataset is more complex compared to the aligned dataset. During the training process, the network for the unaligned dataset experiences greater learning pressure, with heavy consumption of  computing and storage resources, which will challenge the network training to achieve a superior performance.

For the architecture of DeepONet, it is noteworthy to state that the explicit use of two independent subnetworks to handle information with different physical meanings itself is a form of embedded prior knowledge. 
Moreover, the network structures in the branch and trunk net are not restricted and may incorporate a wide range of  structures, like FNNs, CNNs, and RNNs, making it flexible for different problems.

\subsection{POD-DeepONet}\label{pod}

Proper orthogonal decomposition (POD)  is widely used to reduce the dimensionality of  high-dimensional data while preserving the key features of the data\cite{20191}. It projects the high-dimensional data on low-rank basis functions, called POD modes, which characterize the dominant variability in the data. The coefficients of the POD modes represent the amplitude of the modes and can be used to reconstruct the original data.

In equation \eqref{bk}, the output function ${G}(u)(\xi)$ can be seen as a line combination of the branch and trunk net outputs.
To improve the prediction accuracy of DeepONet, \citet{luComprehensiveFairComparison2022} proposed a POD-DeepONet, in which the pre-calculated POD modes are utilized in the trunk net,  the branch net serves to learn the coefficients of the POD modes, and the Cartesian product operation is endowed with a similar operation to the POD reconstruction.

\begin{figure}[h]
	\centering
	\includegraphics[width=1\textwidth,trim=38 240 315 155,clip]{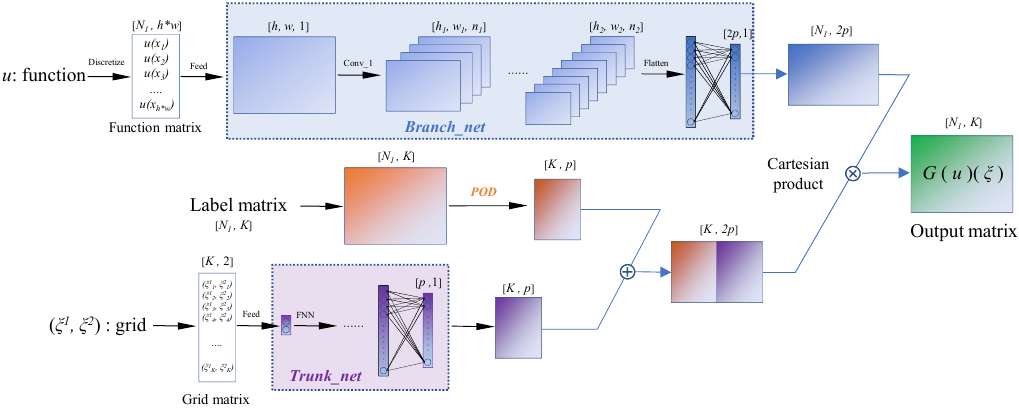}
	\caption{The architecture of POD-DeepONet. Branch net and trunk net are shown in the form of CNNs and FNNs, respectively. Only aligned data can be fed.}
	\label{fig:POD-DeepONet}
\end{figure}

The architecture of the POD-DeepONet is shown in figure \ref{fig:POD-DeepONet}. Specifically, a Matrix splicing operation combines the output of the trunk net and the POD modes, giving a constraint condition that the dimension of the POD modes and the Grid matrix must be kept the same as each other. Although POD-DeepONet exhibits reduced learning pressure and superior accuracy, it can not train the unaligned data, as only one distribution of  observation $\xi$ can be fed into the trunk net.

\subsection{Decoder-DeepONet operator regression }\label{DD}

\begin{figure}[h]
	\centering
	\includegraphics[width=1\textwidth,trim=85 270 130 170,clip]{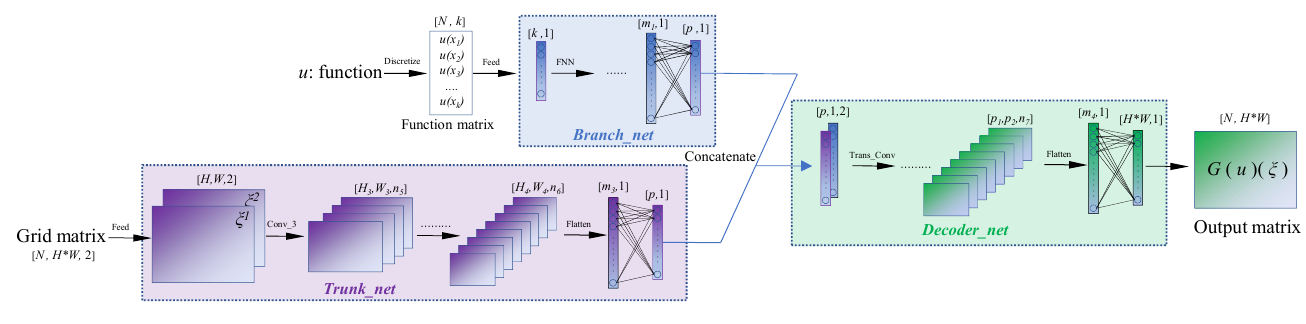}
	\caption{The architecture of hybrid Decoder-DeepOnet operator regression framework. The dimension evolution of unaligned dataset is processed in this figure. Note that H*W in this figure is numerically equal to K in Figure \ref{fig:DeepOnet} and \ref{fig:POD-DeepONet}. We specifically select the use of H*W over K here to illustrate that CNNs are a viable choice if the input data is structured.}
	\label{fig:DeepOnet-AE-2}
\end{figure}

Aiming to improve the prediction accuracy and efficiency of operator regression for unaligned data, we  propose a  hybrid Decoder-DeepONet operator regression framework, which is depicted in figure \ref{fig:DeepOnet-AE-2}.
Given the framework, the hybrid Decoder-DeepONet operator can be expressed as:
\begin{equation}
	G(u)(\xi) \approx \underbrace{\boldsymbol{d}( }_{decoder}\underbrace{\boldsymbol{b}(\textit{u}(x_{1}), \textit{u}(x_{1}), ... ,\textit{u}(x_{k}))}_{branch}, \underbrace{\boldsymbol{t}(\xi)}_{trunk} ), 
	\label{sss}
\end{equation}
in which the explicit use of two independent subnetworks to handle information with different physical meanings is kept, whereas a decoder network $\boldsymbol{d}$ is used to merge the outputs of subnetworks instead of the dot product.
 More importantly, the decoder network allows that the varied observations at different input functions can be fed into the trunk net as a whole distribution, outperforming DeepONet and POD-DeepONet.

We will prove that the hybrid Decoder-DeepONet is also obeying the universal operator approximation theorem, sharing similarities of DeepONet in Theorem 2. 
Under the universal operator approximation theorem, we first prove that the dot product in DeepOnet can be replaced by a decoder network $\boldsymbol{d}$. 
Suppose that $\langle \bm{b}, \bm{t}\rangle: \mathbb{R}^{2p} \to \mathbb{R}$ is the dot product function. The defined domain of $\langle \bm{b}, \bm{t}\rangle$ is $\left\{\bm{b},\bm{t}\right\}$ = $\left\{b_{1}, b_{2}, ..., b_{p}, t_{1}, t_{2}, ..., t_{p}\right\}$, being a compact set. Based on the universal approximation theorem of functional\cite{256500}, the function $\langle \bm{b}, \bm{t}\rangle$ is able to approximated by a decoder network, that is,

\noindent \textbf{Theorem 3}: For any $\epsilon > 0$, there are $x_{j}\in K_{1}$, continuous vector functions
$\bm{b}: \mathbb{R}^{m} \to \mathbb{R}^{p}$, 
$\bm{t}:\mathbb{R}^{d} \to \mathbb{R}^{p}$,
$\bm{d}: \mathbb{R}^{2p} \to \mathbb{R}$, such that:
\begin{equation}
	\left|   \langle \bm{b}(u(x_{1}),u(x_{1}),...,u(x_{m})), \bm{t}(\xi)\rangle - \bm{d}(\bm{b}, \bm{t})\right| < \frac{1}{2}\epsilon, 
	\label{equ:6}
\end{equation}
holds for all $u\in V$ and $\xi \in K_{2}$, where the decoder network $\bm{d}$ needs to satisfy the universal approximation theorem of functional, and all the notations above is shared from Theorem 2.

	Substitute \eqref{equ:6} into \eqref{equ:3}, we get:

\noindent \textbf{Theorem 4}: For any $\epsilon > 0$, there are continuous vector functions
$\bm{b}: \mathbb{R}^{m} \to \mathbb{R}^{p}$, 
$\bm{t}:\mathbb{R}^{d} \to \mathbb{R}^{p}$,
$\bm{d}: \mathbb{R}^{2p} \to \mathbb{R}$, 
such that:
\begin{equation}
	\left| G(u)(\xi) - \bm{d}(\bm{b}(u(x_{1}),u(x_{1}),...,u(x_{m})), \bm{t}(\xi))\right| < \epsilon, 
\end{equation}
holds for all $u\in V$ and $\xi \in K_{2}$.

Theorem 4 states that a decoder can serve as a substitute for the dot product operation utilized in DeepOnet. This operation will offer several advantages over DeepOnet:

(i) As an additional subnetwork, the decoder enhances the expressive power of the entire learning framework, enabling it to provide multidimensional outputs and facilitating efficient processing of unaligned data.

(ii) Regarding the dimension evolution of unaligned dataset in figure \ref{fig:DeepOnet-AE-2}, the function matrix and grid matrix have the dimensions of $[N, k]$ and $[N, K, 2]$ to be fed into the branch and trunk net, respectively.
In contrast to DeepONet, which employs 2 neurons for receiving a single 2D observation to correspond with each function, Decoder-DeepOnet accommodates the $K$ of 2D observations to match every function by utilizing $K*2$ neurons, which significantly improves the efficiency of processing unaligned data.
In figure \ref{fig:DeepOnet-AE-2}, the $K$ observations can be reshaped into a structured shape of $H\times W$, offering a good choice of CNNs to learn the high-dimensional structured  data.

(iii) Additionally, the outputs of the branch nets and trunk net are combined through splicing into different channels, which is the input of the decoder net to approximate the final output with the $K$ neurons in the last layer.
Notably, besides the splicing operation to concatenate the outputs of subnetworks, addition and multiplication operation can also be utilized. Considering that the splicing operation preserves more dimensional and positional information and enables the subsequent layers of the decoder to choose features flexibly\cite{10.1007/978-3-319-24574-4_28}, we use it 
straightforwardly in the hybrid network.

When processing unaligned dataset as shown in figure \ref{fig:DeepOnet-AE-2}, different from DeepOnet in which  $\xi \in K_{2} \subset \mathbb{R}^{d}$, $\xi$ in the grid matrix of Decoder-DeepONet can be a vector of $K\cdot d$ dimensions in $\mathbb{R}^{K\cdot d}$ ($K$=$H*W$, $d$=2 in figure \ref{fig:DeepOnet-AE-2}). Similarly, $G$ in the output matrix of Decoder-DeepONet is a vector of $K$ dimensions. 
It can be considered as a special case of Theorem 4 in which we feed the observation coordinate $\xi$ with dimensions of $h\cdot d$ and get the output $G$ with dimensions of $h$. So Theorem 4 still holds with $\bm{t}:\mathbb{R}^{h*d} \to \mathbb{R}^{p}$ and $\bm{d}: \mathbb{R}^{2p} \to \mathbb{R}^{h}$.

For the loss function, the mean squared error (MSE) is adopt to measure the square of L2 norm between predictions and true values. Given the data size in figure \ref{fig:DeepOnet-AE-2}, the loss is expressed as:
\begin{equation}
    \mathcal{L} = \frac{1}{N\cdot H\cdot W}\sum\limits_{i=1}^{N}\sum\limits_{j=1}^{H\cdot W}(G_{model}(\textit{u}_{i})(\xi_{j})-G_{truth}(\textit{u}_{i})(\xi_{j}))^{2},
\end{equation}
where $G_{truth}(\textit{u}_{i})(\xi_{j})$ means the ground-truth data under the i-th input function at j-th observation point and $G_{truth}(\textit{u}_{i})(\xi_{j})$ is the corresponding output of the model learnt by Decoder-DeepOnet.

\subsection{Multi-Decoder-DeepOnet with input augmentation}\label{MDD}

Drawing inspiration from the multiple-input operator\cite{jinMIONetLearningMultipleInput2022a}, Decoder-DeepOnet can be easily extended to a multiple-input form. In this study, we provide a Multi-Decoder-DeepOnet, which incorporates an input augmentation to further improve the prediction accuracy of operator regression.

Fruitful prior knowledge is hiding in the label data. Here we utilize the mean field of the label data as an input augmentation to incorporate the information of the known label data.
The mean field of the label data is expressed as,
 \begin{equation}
 	\overline{G} =\frac{1}{N} \sum\limits_{i=1}^{N}G_{i}(u_{i})(\xi),
 \end{equation}
where $N$ is the number of the output functions in training data. 

\begin{figure}[h]
	\centering
	\includegraphics[width=1\textwidth,trim=60 210 140 160,clip]{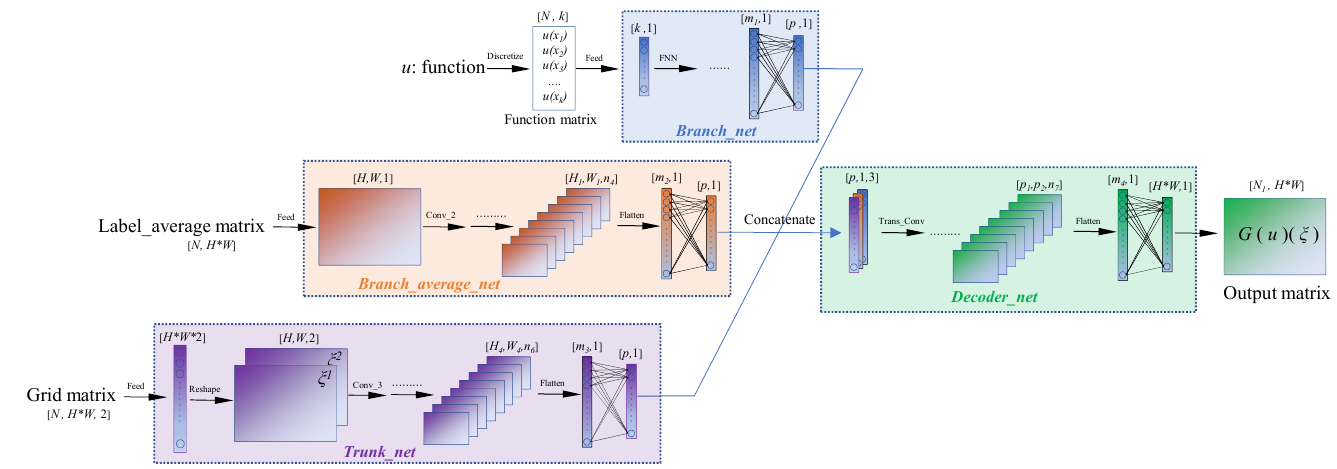}
	\caption{The architecture of hybrid Multi-Decoder-DeepOnet operator regression framework. 
	}
	\label{fig:DeepOnet-AE}
\end{figure}

The architecture of the Multi-Decoder-DeepOnet is shown in figure \ref{fig:DeepOnet-AE}.
A branch-average-net is added to receive the mean field of the label data. The branch-average-net is responsible for extracting  features in the mean field, providing a warm start for operator learning. 
Owning to the decoder net's flexibility, the information in the branch-average-net can be integrated and decoded to generate the corresponding field data.


Given the framework in figure \ref{fig:DeepOnet-AE}, the Multi-Decoder-DeepONet can be expressed as:
\begin{equation}
	G(u)(\overline{G})(\xi) \approx \underbrace{\boldsymbol{d} ( }_{decoder}\underbrace{\boldsymbol{b}(\textit{u}(x_{1}), \textit{u}(x_{1}), ... ,\textit{u}(x_{k}))}_{branch},\underbrace {\boldsymbol{b}_{a}(\overline{G})}_{branch_{a}}, \underbrace{t(\xi)}_{trunk} ) 
	\label{sss},
\end{equation}
where $\boldsymbol{b}_{a}$ represents the branch-average-net.

The approximation theory of Multi-Decoder-DeepONet with input augmentation is given as,

\noindent \textbf{Theorem 5}: For any $\epsilon > 0$, there are $x_{j}\in K_{1}$, 
$\bm{b_{1}}: \mathbb{R}^{m} \to \mathbb{R}^{p}$, 
$\bm{b_{2}}: \mathbb{R}^{h} \to \mathbb{R}^{p}$, 
$\bm{f}:\mathbb{R}^{h \cdot d} \to \mathbb{R}^{p}$,
$\bm{d}: \mathbb{R}^{3p} \to \mathbb{R}^{h}$, 
such that :
\begin{equation}
	\left| G(u)(\overline{G})(\xi) - \bm{d}(\bm{b_{1}}(u(x_{1}),u(x_{1}),...,u(x_{m})), \bm{b_{2}}(\overline{G}), \bm{t}(\xi))\right| < \epsilon, 
\end{equation}
holds for all $u\in V$ and $\xi \in K_{2}$. The proof is referred to the study of \citet{jinMIONetLearningMultipleInput2022a}, in which the approximation theory for multiple-input operator regression was first proposed.


Similar to to Decoder-DeepONet, the loss in the training of Multi-Decoder-DeepONet is given as, 
\begin{equation}
    \mathcal{L} = \frac{1}{N*H*W}\sum\limits_{i=1}^{N}\sum\limits_{j=1}^{H*W}(G_{model}(\textit{u}_{i})(\overline{G})(\xi_{j})-G_{truth}(\textit{u}_{i})(\overline{G})(\xi_{j}))^{2}.
\end{equation}

\section{Numerical experiments}\label{results}
Two numerical experiments are provided here to demonstrate the efficiency and accuracy of the proposed methods. 
The network training is performed on NVIDIA GeForce RTX 3090 GPU. All codes and data can be found in GitHub at \href{https://github.com/cb-sjtu/Decoder_DeepONet.git}{github\_Deocer\_DeepONet}, which are developed on the basis of DEEPXDE\cite{lu2021deepxde}, a library for scientific machine learning, with Tensorflow as the backend.

\subsection{Two-dimensional Darcy flows}
The Darcy flow is a benchmark problem  for operator learning. In previous studies\cite{li_fourier_2021,luComprehensiveFairComparison2022}, only aligned datasets of the Darcy problem have been tested. Here we focus on the superiority of Decoder-DeepOnet and Multi-Decoder-DeepOnet for unaligned data, in terms of learning accuracy and memory usage, compared to the basic DeepONet.

\subsubsection{Problem setup}

The governing equations of two-dimensional Darcy flows are  linear elliptic PDEs\cite{luComprehensiveFairComparison2022},
 \begin{equation}
 \left\{
     \begin{array}{cc}
          -\bigtriangledown \cdot (k(\xi^{1},\xi^{2})\bigtriangledown h(\xi^{1},\xi^{2})) = f,\\
          h(\xi^{1},\xi^{2}) = 0,
     \end{array}
\right.
 \end{equation}
where  $(\xi^{1},\xi^{2})$ is the spatial coordinates, $k$ and $h$ denote the permeability and pressure field, respectively, and $f$ is a forcing function which can be either a constant or a function. Here the forcing is fixed as $f$ = 1.

 The permeability field denoted by $k$ is mathematically defined as $k=\psi(\mu)$, where the random variable $\mu$ corresponds to the Gaussian random field with zero Neumann boundary conditions on the Laplacian and is evaluated at points $\mu=R(0,(-\bigtriangleup+9I)^{-2})$. The push-forward of the mapping $\psi$ is defined pointwise, with a value of 12 assigned to the positive segment of the real line and 3 assigned to its negative counterpart. 
 
The target is to learn the operator mapping from the permeability field $K(\xi)$ to the pressure field $h(k)(\xi)$:
 \begin{equation}
     \mathcal{G}: K(\xi) \mapsto h(k)(\xi).
 \end{equation}
 
 \subsubsection{Data generation}
 
A rectangular domain is set with  $(\xi^{1},\xi^{2})\in{(0,1)}$.
 The data set for training and testing is generated using the MATLAB Partial Differential Equation Toolbox\cite{li_fourier_2021}.
 Notably, the grids $\xi$ for each different $h(k)$ is sampled randomly via different seeds to generate unaligned data like Figure \ref{fig:Darcy}. Three kinds of cases with different resolutions ($29\times29, 43\times43, 61\times61$) are used here to eliminate the impact of sampling resolution.
 
In the training process, $k(\xi)$ and the averaged pressure filed $\overline{h}$ are fed into the branch net and branch-average-net, respectively. The trunk net receives the gird distributions. Finally, $h(\xi)$ is operated as the ground-truth data to estimate the loss of the operator approximation, which is defined in section \ref{deeponet}. In general, 1000 samples are selected randomly for training, and 200 samples are used to evaluate the capabilities of the trained models concerning generalization.

 Table \ref{tab:1} shows the data size comparison between DeepONet and Decoder-DeepOnet for unaligned data. It is evident that for identical resolution data, the Decoder-DeepOnet dataset encapsulates training information more efficiently than DeepONet. The data redundancy in DeepONet is primarily attributable to the function matrix, in which the same function repeatedly matches each single observation point. It should be noted that the data of Multi-Decoder-DeepOnet includes only one additional label-average matrix of identical size to the function matrix of Decoder-DeepOnet, compared to Decoder-DeepOnet data. As POD-DeepONet is incapable of managing unaligned data, it is omitted for comparison.
 
  \begin{table}[h]
 	\centering
 	\caption{Training data size of DeepONet and Multi-Decoder-DeepOnet for the unaligned Darcy problem }
 	\resizebox{\linewidth}{!}{
 		\begin{tabular}{ccccccc}
 			\toprule 
 			Case & Method  &  Resolution of $k$ & Function matrix& Gird matrix& Output matrix  \\ 
 			\midrule 
 			case-1 & DeepONet & 29*29 & [1000*841, 841] & [1000*841,2]&[1000*841,1]\\
 			case-2 & Decoder-DeepOnet & 29*29&[1000, 841] & [1000,841,2]&[1000,841] \\
 			case-3 & DeepONet & 43*43 & [1000*1849, 1849] & [1000*1849,2]&[1000*1849,1]\\
 			case-4 & Decoder-DeepOnet & 43*43&[1000, 1849] & [1000,1849,2]&[1000,1849] \\
 			case-5 & DeepOnet & 61*61& [1000*3721,841]	& [1000*3721,2]&[1000*3721,1] \\
 			case-6 & Decoder-DeepOnet & 61*61&[1000, 841] & [1000,3721,2]&[1000,3721] \\
 			\bottomrule 
 		\end{tabular}
 	}
 	\label{tab:1}
 \end{table}
 
 \subsubsection{Results}
 As the input datasets of the subnetworks exhibit structured characteristics, we employ CNNs, incorporating pooling layers and a fully connected layer in the final stage, to facilitate superior feature extraction from the permeability field, the average pressure field and the grid distribution. FNNs are employed in the decoder net here.
 
The specific hyperparameters of Decoder-DeepOnet and Multi-Decoder-DeepOnet are displayed in \ref{ApDAR}. The sets of DeepONet are the same as Ref. \cite{luLearningNonlinearOperators2021}. We train the models using the Adam optimizer\cite{kingmaAdamMethodStochastic2017} with a learning rate of 1e-5. 

The training and testing  errors are listed in Table \ref{tab:2}. 
It can be seen that, regardless of the grid resolutions, Decoder-DeepONet performs better than DeepONet in reducing the training and testing errors for the unaligned data, and superior performance is achieved by Multi-Decoder-DeepONet with input augmentation.

Given the enhanced descriptiveness of the output function with increased observations, the impact of observation distribution is mitigated for high-resolution data compared to low-resolution data. 
This assertion is corroborated by the data presented in Table \ref{tab:2}, which illustrates that superior learning performance is consistently achieved at higher resolutions, irrespective of the methodology employed, and the improvement in learning accuracy brought about by Decoder-DeepONet and Multi-Decoder-DeepONet is markedly significant at lower resolutions.

It is pertinent to note, however, that the observational distribution utilized in the experiments is randomly generated, 
devoid of physical significance or systematic patterns, thus rendering the trunk net incapable of effectively mapping the observation input to a low-dimensional latent space. 
Consequently, despite the substantial enhancements achieved with the Decoder-DeepONet and Multi-Decoder-DeepONet, prediction accuracy remains somewhat limited. 
To further illustrate the superiority of the proposed frameworks, the following experiment will explore a natural unaligned airfoil flow field problem.
 \begin{table}[h]
     \centering
     \caption{MSE loss of DeepONet, Decoder-DeepONet and Multi-Decoder-DeepONet for the Darcy problem }
     \begin{tabular}{ccccc}
     \toprule 
     case & method  &  resolution & training error& testing error \\ 
     \midrule 
     case-1 & DeepONet & 29*29 & 2.71e-02 & 3.61e-02\\
     case-2 & Decoder-DeepOnet & 29*29&4.35e-03 & 1.23e-02 \\
     case-3 & Multi-Decoder-DeepOnet & 29*29&5.53e-03 & 8.98e-03 \\
     case-4 & DeepONet & 43*43 & 2.95e-02 & 3.62e-02\\
     case-5 & Decoder-DeepOnet & 43*43 & 3.14e-03 & 1.03e-02\\
     case-6 & Multi-Decoder-DeepOnet & 43*43&3.95e-03 & 7.73e-03 \\
     case-7 & DeepOnet & 61*61& 2.76e-02	& 2.33e-02 \\
     case-8 & Decoder-DeepOnet & 61*61& 3.13e-03	& 7.48e-03 \\
     case-9 & Multi-Decoder-DeepOnet & 61*61 & 2.17e-03 & 6.53e-03 \\
     \bottomrule 
     \end{tabular}
     
     \label{tab:2}
 \end{table}



\subsection{Prediction of the flow-field around an airfoil}

The prediction of the flow-field around an airfoil is always achieved by solving Navier-Stokes equations, which will consume high computing  cost with traditional computational fluid dynamics (CFD) methods\cite{duAirfoilDesignSurrogate2022}.

Alternatively, fast surrogate models with DNNs have been used for the flow-field predictions\cite{bhatnagarPredictionAerodynamicFlow2019,duRapidAirfoilDesign2021a,duruDeepLearningApproach2022,huNeuralNetworksBasedAerodynamic2020,maPressureDistributionPrediction2022,zuoFastSparseFlow2022}. 
 Different airfoil shapes (wall boundary conditions) corresponds to different the flow-fields around  the airfoils (observations). It is a natural unaligned problem as the observation distributions (computational grids) vary in accordance with the airfoil shapes.
 Most previous studies did not account for the influence of unaligned data (mesh non-uniformity).
 Recently the learning of operators makes up the deficiencies\cite{huMeshConvConvolutionOperator2022,shuklaDeepNeuralOperators2023}, but still face challenges in network training.
 
Here we use this unaligned problem to demonstrate the capability and efficiency of Decoder-DeepONet and Multi-Decoder-DeepONet in learning operators of unaligned data compared with DeepONet. Additionally, to underscore the significance of accommodating unaligned observations, we also conduct experiments using DeepONet and POD-DeepONet methods to learn the unaligned data in an aligned manner, namely, by feeding only a uniform observation distribution into the trunk net. The specific problem set up and data generation are described below.
\subsubsection{Problem setup}

Focusing on the pressure distributions around  the airfoil, the ground truth data for training and testing is generated from the simulation of compressible Reynolds-averaged Navier-Stokes (RANS) equations. We consider the flow condition: Reynolds number  $Re=3\times{10^6}$, the incoming flow Mach number $Ma=0.5$ and the angle of attack being  $2.5^{\circ}$. 


\begin{figure}[h]
\centering
\includegraphics[width=0.9\textwidth,trim=50 50 290 80,clip]{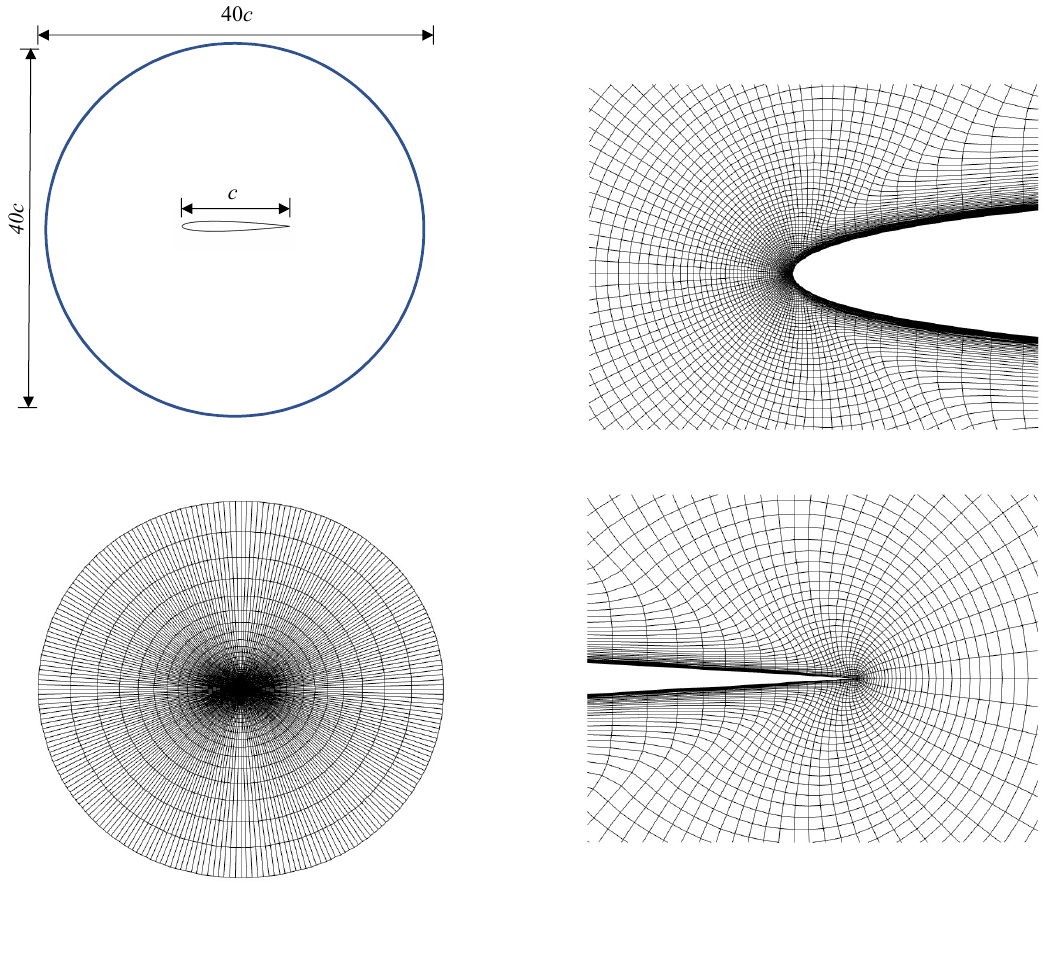}
\caption{\label{fig:mesh} The computational grids. }
\end{figure}
  
\subsubsection{Data generation}

 The CFD is performed with an open source code $CFL3D$ \cite{kristCFL3DUserManual}, and all solutions are calculated using k-$\omega$ SST two equation turbulence model. The validation is provided in \ref{ApA}. 
 A shape space containing 200 airfoils, which varies from NACA0005 to NACA5535, is generated. 
 The O-shaped structured grids shown in figure \ref{fig:mesh} are automatically generated by using scripts of pyGlyph based on python. To eliminate the influence of boundaries on the flow field, we set the chord length of the O-type computational domain to $20c$ and $c$ is the chord length of airfoils. The grid matrix shape of the computational domain is $241\times120$ and a $241\times100$ internal area is selected for the network training. 
 
  In the training process, the airfoil shape is represented by 256 grid points and fed into the branch net directly without independent parameter dimensionality reduction. The branch net receives the airfoil shapes as the wall boundary conditions, and the $241\times100$ mesh grids are fed into the trunk net. The pressure distributions obtained by CFD method are regarded as the ground-truth data to examine the loss of the operator approximation, which is defined in section \ref{method}. 
   \begin{table}[h]
 	\centering
 	\small
 	\caption{MSE loss of the flow-filed around the airfoil }
 	\begin{tabular}{ccccc}
 		\toprule 
 		case & method  &  resolution & training error& testing error \\ 
 		\midrule 
 		case-1 & DeepONet & 121*100 & 5.87e-04 & 1.83e-02\\
 		case-2 & DeepONet (aligned method) & 121*100 & 5.53e-05 & 8.44e-05\\
 		case-3 & POD-DeepONet (aligned method) & 121*100 & 8.20e-07 & 4.49e-07\\
 		case-4 & Decoder-DeepOnet & 121*100&3.60e-07 & 4.11e-07 \\
 		case-5 & Multi-Decoder-DeepOnet & 121*100&5.43e-08 & 1.67e-07 \\
 		\bottomrule 
 	\end{tabular}
 	
 	\label{tab:3}
 \end{table}

  \begin{figure}[!h]
 	\centering 
 	\subfigure
 	{\includegraphics[width=0.4\textwidth,trim=10 10 10 10,clip]{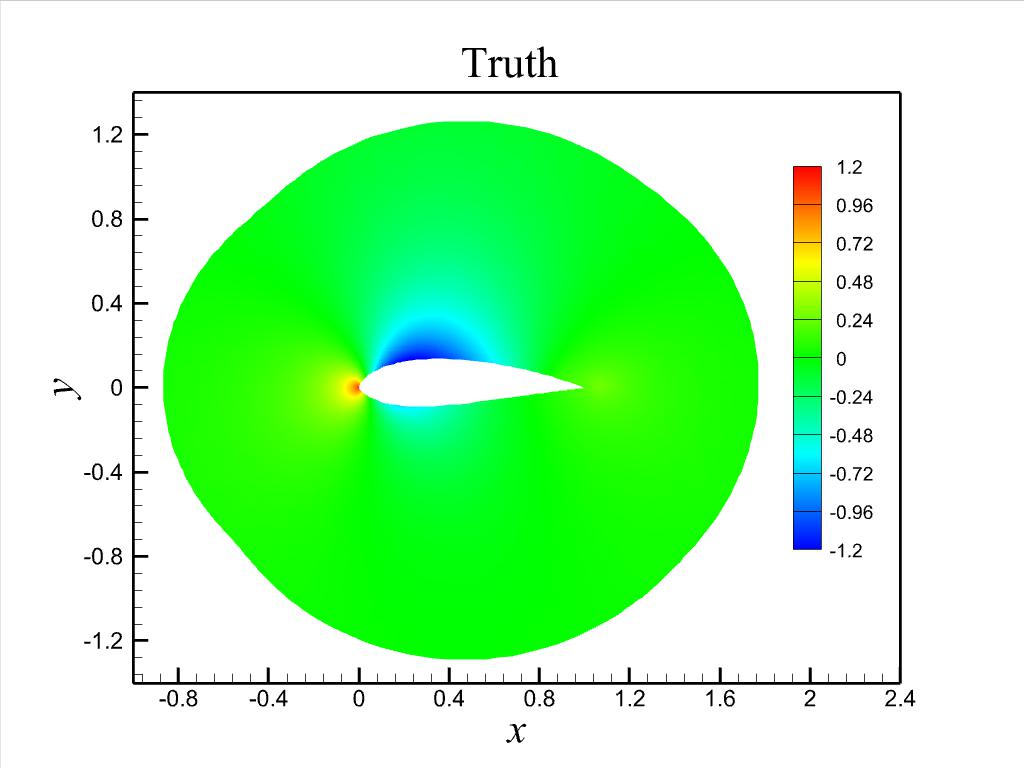}}
 	\subfigure
 	{\includegraphics[width=0.4\textwidth,trim=10 10 10 10,clip]{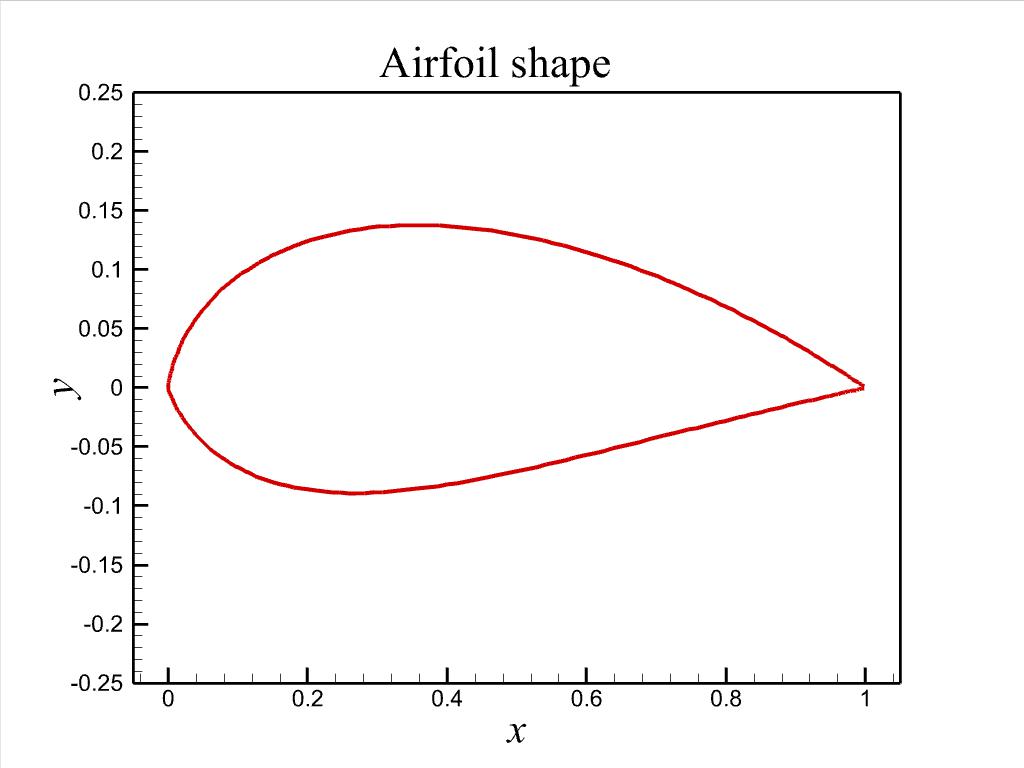}}
 	\subfigure
 	{\includegraphics[width=0.4\textwidth,trim=10 10 10 10,clip]{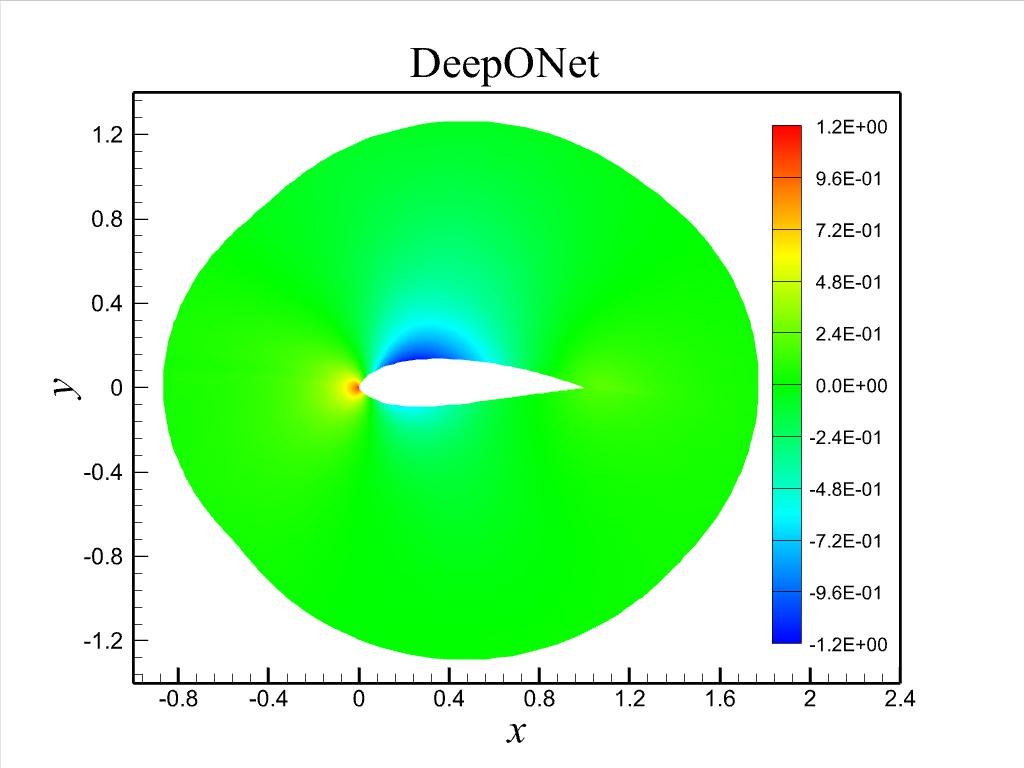}}
 	\subfigure
 	{\includegraphics[width=0.4\textwidth,trim=10 10 10 10,clip]{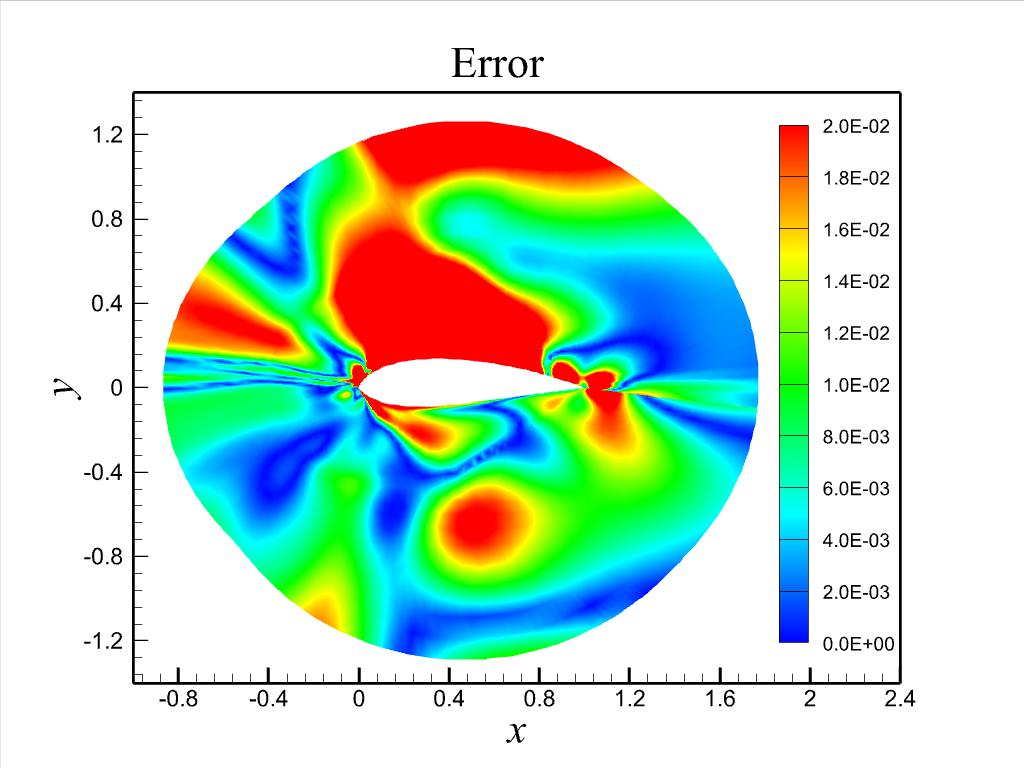}}
 	\subfigure
 	{\includegraphics[width=0.4\textwidth,trim=10 10 10 10,clip]{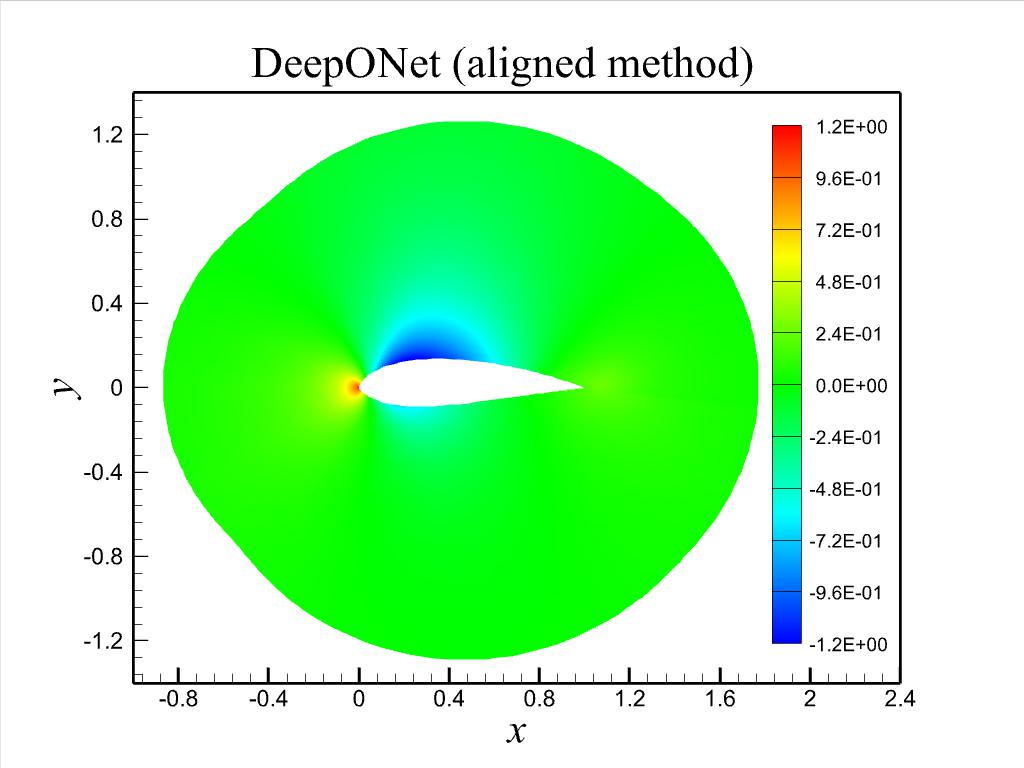}}
 	\subfigure
 	{\includegraphics[width=0.4\textwidth,trim=10 10 10 10,clip]{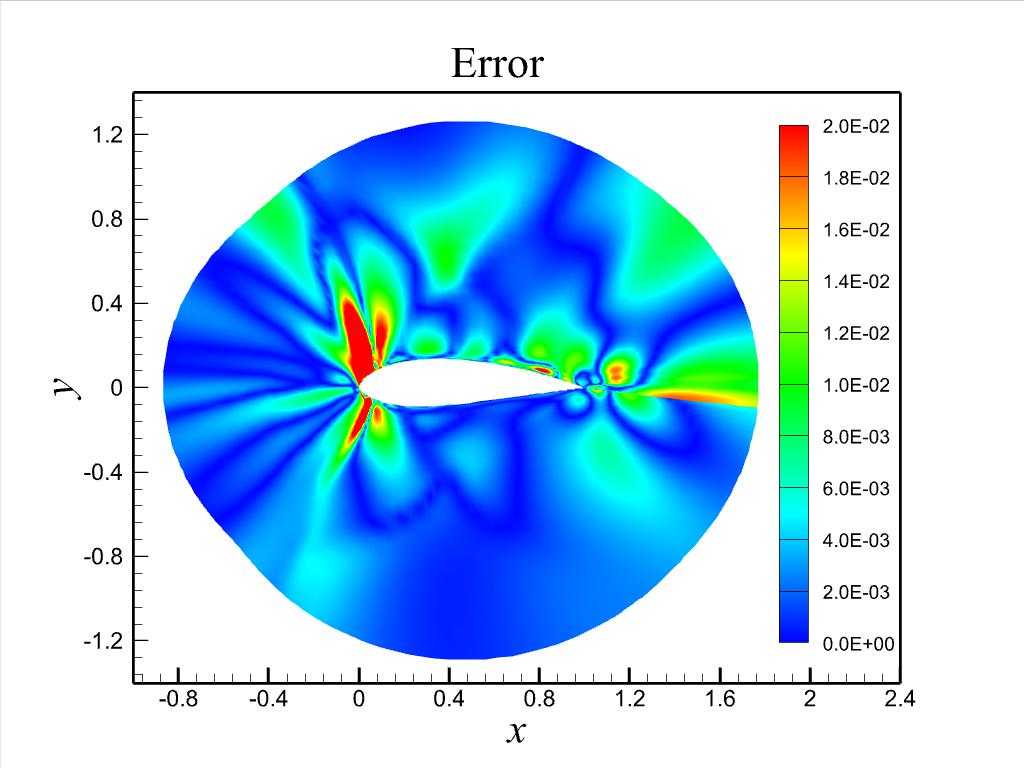}}
 	\subfigure
 	{\includegraphics[width=0.4\textwidth,trim=10 10 10 10,clip]{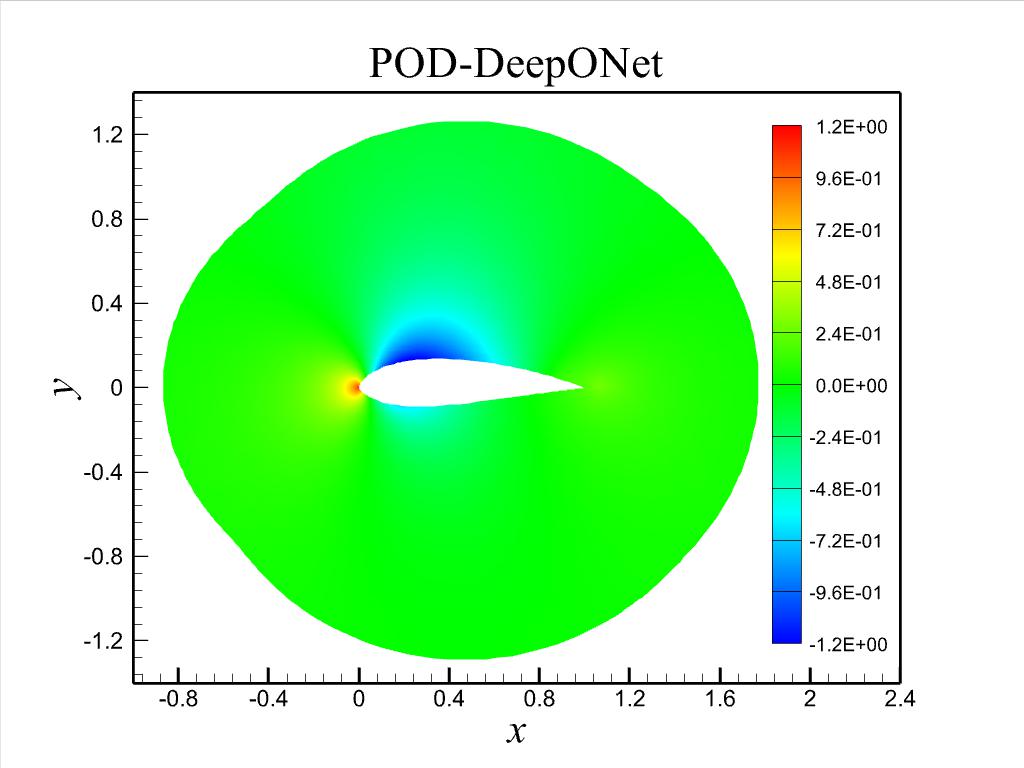}}
 	\subfigure
 	{\includegraphics[width=0.4\textwidth,trim=10 10 10 10,clip]{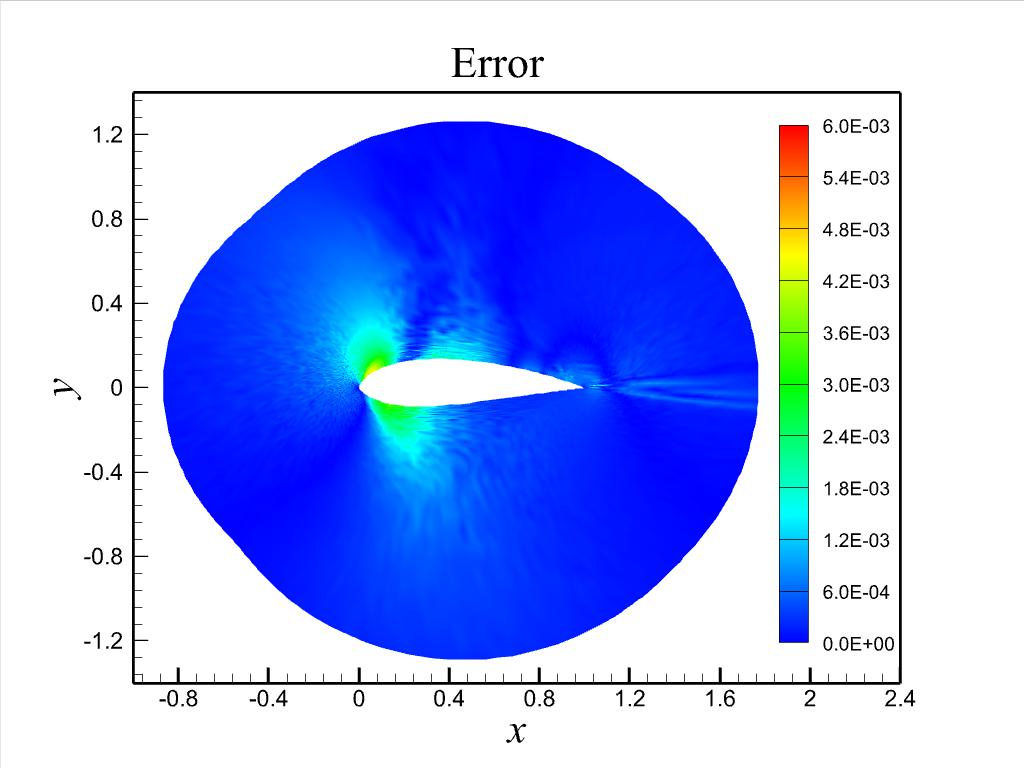}}
 	\label{fig:Airfoil-1}
 \end{figure}
 
 \begin{figure}[!h]
 	\centering 	
 	\subfigure
 	{\includegraphics[width=0.4\textwidth,trim=10 10 10 10,clip]{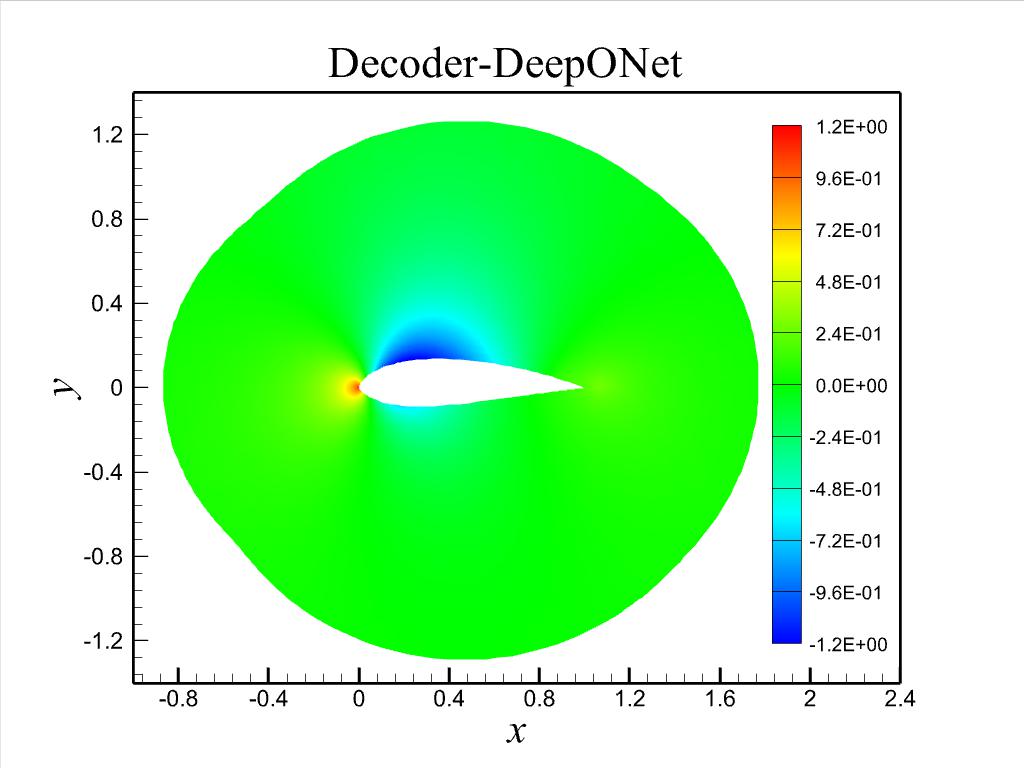}}
 	\subfigure
 	{\includegraphics[width=0.4\textwidth,trim=10 10 10 10,clip]{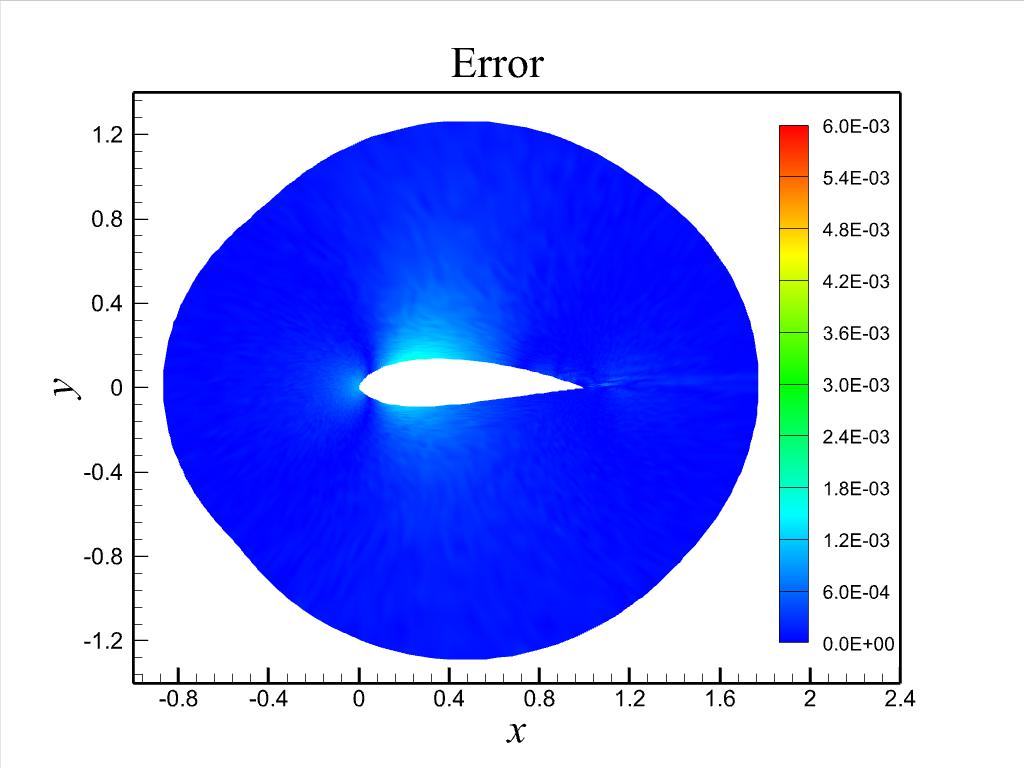}}
 	\subfigure
 	{\includegraphics[width=0.4\textwidth,trim=10 10 10 10,clip]{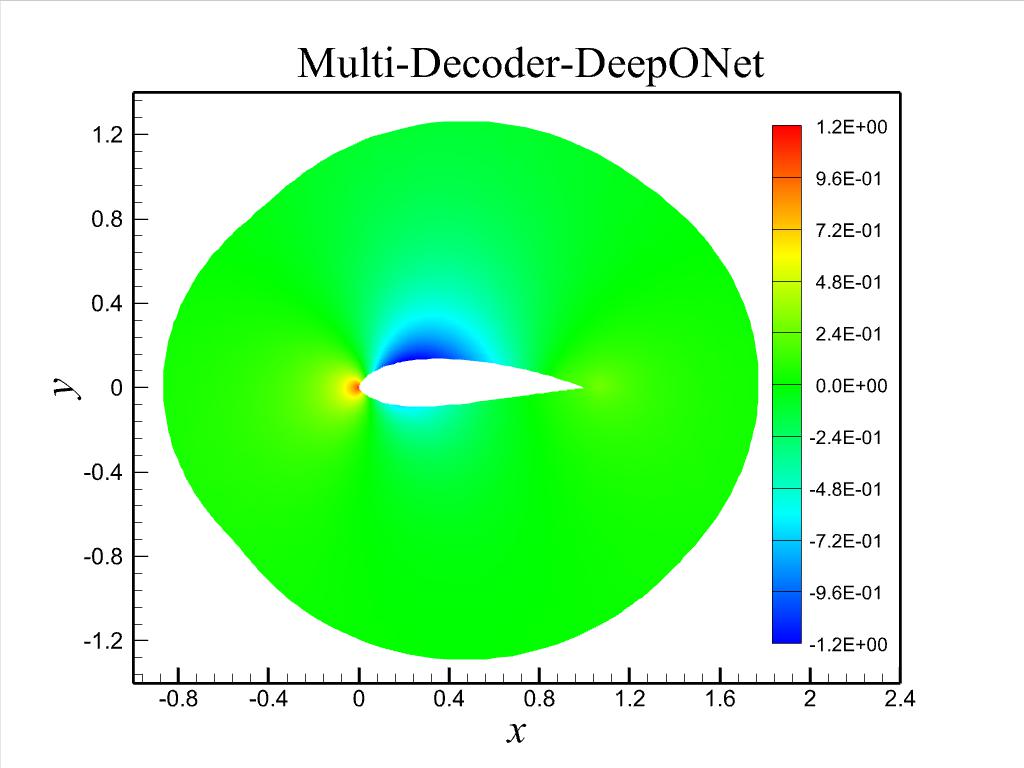}}
 	\subfigure
 	{\includegraphics[width=0.4\textwidth,trim=10 10 10 10,clip]{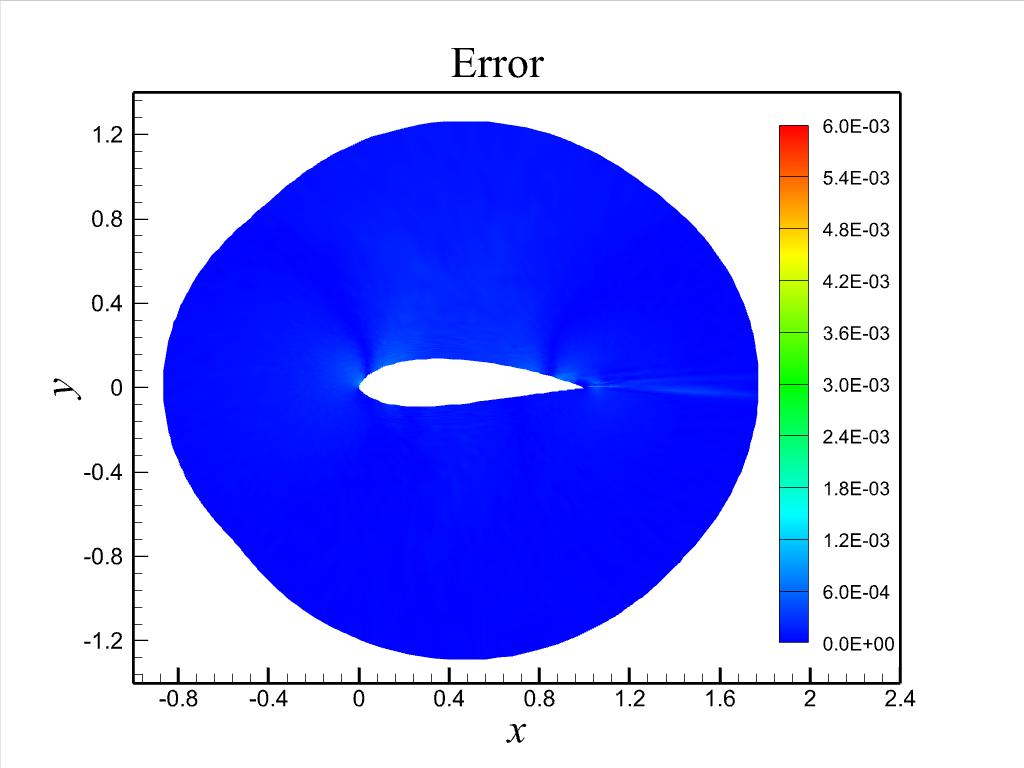}}
 	\caption{The predicted pressure coefficients and the absolute errors around a randomly selected airfoil from the testing set}
 	\label{fig:Airfoil-2}
 \end{figure}

 \subsubsection{Results}

  We compare Decoder-DeepOnet and Multi-Decoder-DeepOnet with  other three classical operator methods: DeepONet with real unaligned observations, DeepONet with aligned uniform observations and POD-DeepONet with aligned uniform observations.
Details of the hyperparameters are provided in  \ref{ApAIR}. We train  the models using the Adam optimizer\cite{kingmaAdamMethodStochastic2017} with a learning rate of 1e-4. The resultant mean squared errors, during training and testing, are listed in Table \ref{tab:3}. 

    \begin{figure}[!h]
 	\centering 
 	\subfigure
 	{\includegraphics[width=0.4\textwidth,trim=80 170 210 40,clip]{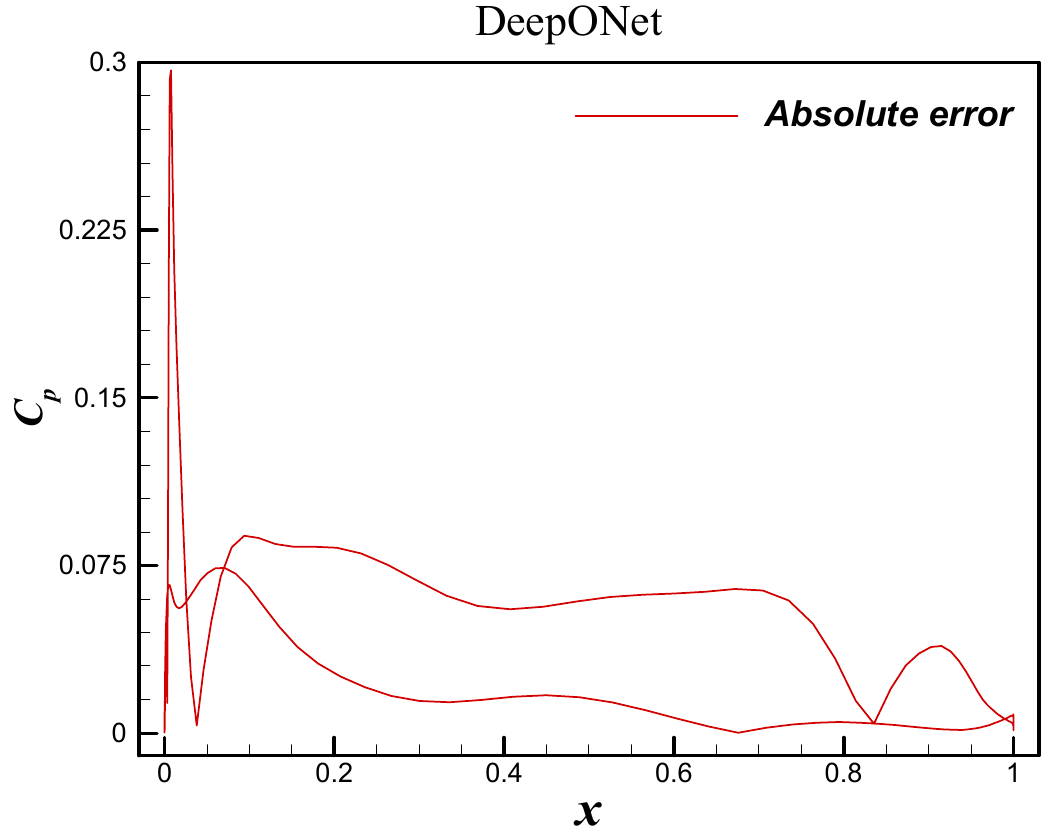}}
 	\subfigure
 	{\includegraphics[width=0.4\textwidth,trim=80 170 210 40,clip]{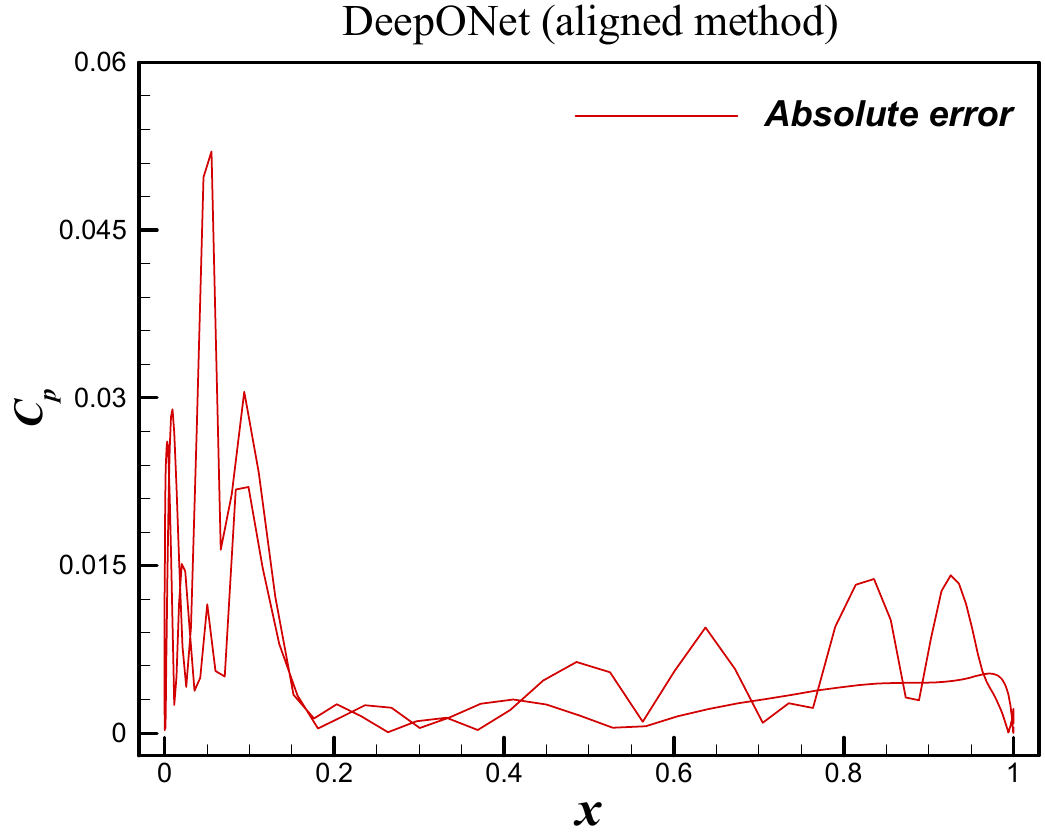}}
 	\subfigure
 	{\includegraphics[width=0.4\textwidth,trim=80 170 210 40,clip]{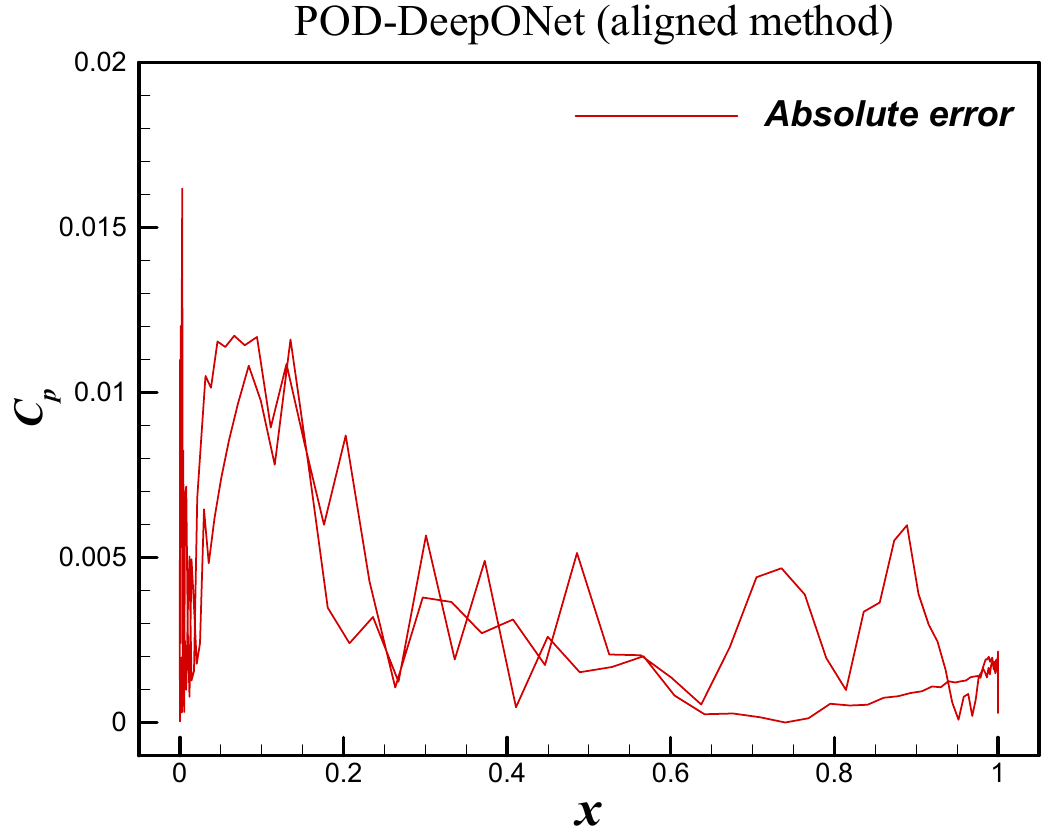}}
 	\subfigure
 	{\includegraphics[width=0.4\textwidth,trim=80 170 210 40,clip]{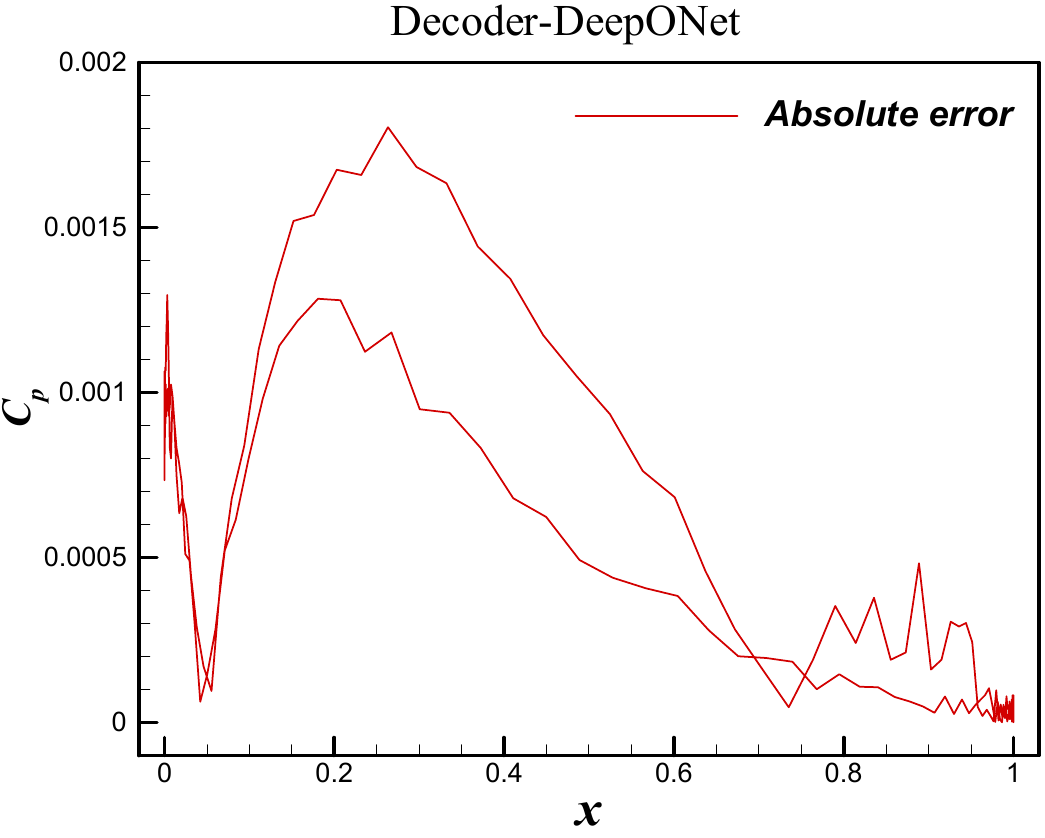}}
 	\subfigure
 	{\includegraphics[width=0.4\textwidth,trim=80 170 210 40,clip]{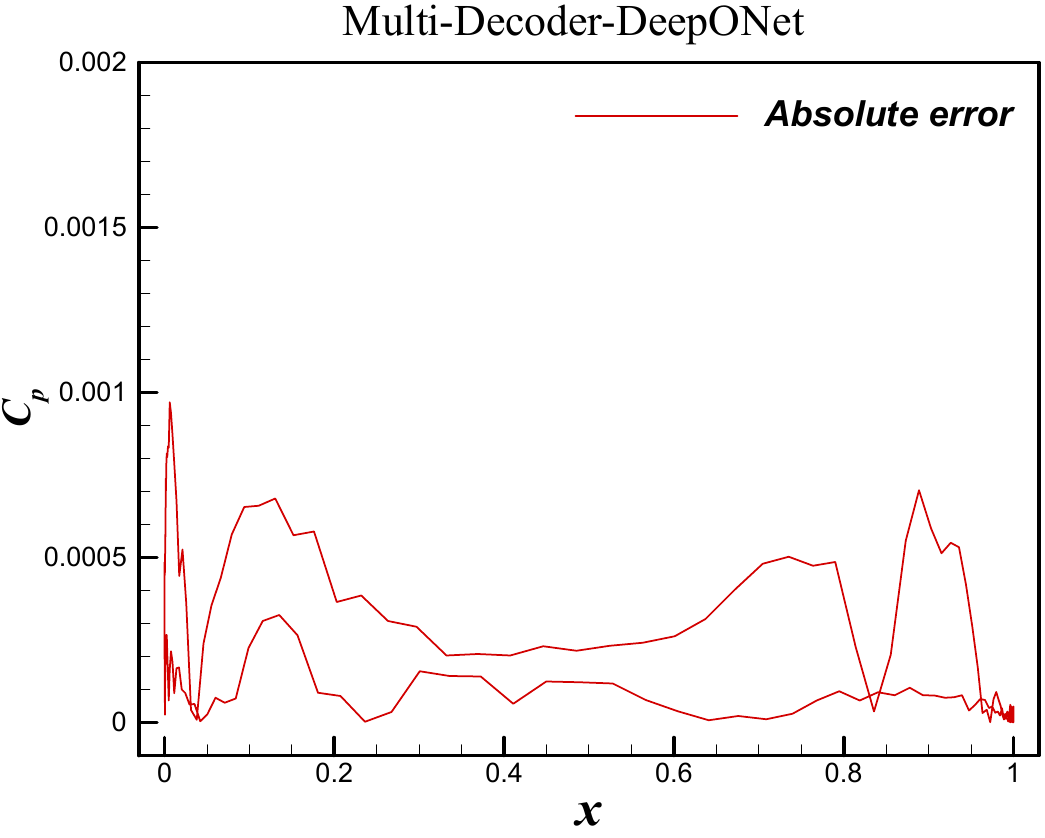}}
 	\caption{The absolute errors between the truth and prediction of pressure coefficients distribution on the airfoil surface.}
 	\label{fig:wall_cp}
 \end{figure}
 

 \begin{figure}[h]
 	\centering
 	\includegraphics[width=0.9\textwidth]{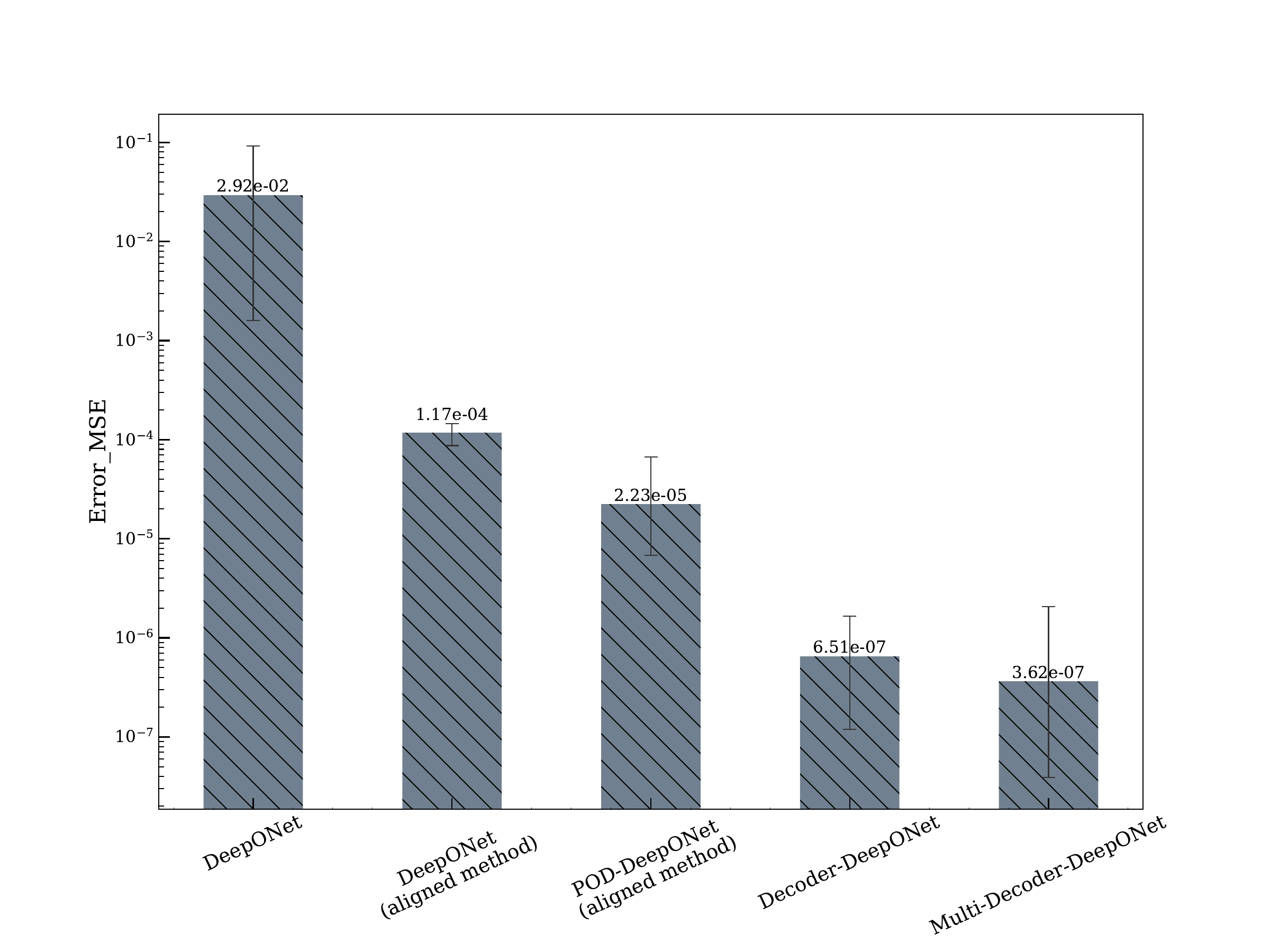}
 	\caption{The MSE errors of all testing cases on the airfoils surface.}
 	\label{fig:wall_mse}
 \end{figure}

First, the performance of DeepONet using unaligned observations exhibits suboptimal results.
As the closest comparison, DeepONet utilizing aligned observations shows lower errors, particularly in terms of testing errors.
The underperformance of the former can be attributed to the more intricate mapping relationship to be learned and the vast quantity of training data employed. 
However, expanding the network structure in terms of depth and width does not alleviate this situation.
Implementation of the POD-DeepONet leads to reduced learning pressure and improved accuracy. Although  the POD-DeepONet and DeepONet with aligned observations benefit from a simpler mapping relationship and smaller training dataset size, it is not physically justifiable to consider unaligned data as aligned.

As for Decoder-DeepONet and Multi-Decoder-DeepONet, the prediction errors are several orders of magnitude lower than other methods,  and Multi-Decoder-DeepONet with input augmentation delivers the most superior performance. 
With receiving real unaligned observations, Decoder-DeepONet and Multi-Decoder-DeepONet process data in a more efficient way without a significant increase of the training data size. 

Figure \ref{fig:Airfoil-2} illustrates the predicted pressure coefficients  and the absolute errors around a randomly selected airfoil from the testing set. It can be seen that Decoder-DeepONet and Multi-Decoder-DeepONet substantially outperform the other methods in predicting the flow field. Notably, the flow field predicted by Multi-Decoder-DeepONet aligns well with the true flow field.
In addition, the prediction errors are predominantly localized near the wall surface, where fine grid distribution and considerable gradients of pressure change impact the aerodynamic performance of the airfoil. 

Figure \ref{fig:wall_cp} shows the absolute errors between the truth and predicted pressure coefficients on the wall surface. The predictions of DeepONet using unaligned method have the worst agreement with the truth values, as expected. Reductions of the errors are observed for DeepONet and POD-DeepONet using the alignment method. In contrast, Decoder-DeepONet and Multi-Decoder-DeepONet reduce the errors significantly, especially for Multi-Decoder-DeepONet. 
Figure \ref{fig:wall_mse}  shows the MSE errors of pressure coefficients on the wall surface for all test cases, quantitively illustrate the advantages of Decoder-DeepONet and Multi-Decoder-DeepONet over the other methods.

In summary, incorporating decoder nets, Decoder-DeepONet  receives physical grid coordinates in the trunk net, which involving the non-uniformity of the grid in the near-wall region. Multi-Decoder-DeepONet further includes the averaged flow field as an input enhancement. Results demonstrate the superiority of the new methods in handling the unaligned observation data.


\section{Conclusions}\label{conclusion}
In this study, we propose a hybrid Decoder-DeepONet operator regression framework, and extend it to the Multi-Decoder-DeepONet by incorporating an input augmentation. 
Two numerical experiments are performed to evaluate the efficiency and accuracy of the proposed methods.
Conclusions are summarized below:

(i) Replacing the dot product by a decoder net, Decoder-DeepONet still obeys the operator approximation theory. It can be easily extended to an input-augmentation form, due to the flexibility of the decoder net. We give Multi-Decoder-DeepONet, which utilizes the mean field of the label data as the input augmentation.

(ii) In comparison with the classical forms of DeepONet, the proposed methods are validated to be more efficient and accurate in the operator regression of unaligned data problems. The improvements may contribute to a broader understanding of operator approximation and  accelerate the practical applications in various fields. 





\section*{Declaration of competing interest}
The authors declare that they have no known competing financial interests or personal relationships that could have appeared to influence the work reported in this paper.

\section*{Declaration of generative AI in scientific writing}
During the preparation of this work the authors used Chatgpt4 in order to improve readability. After using this tool/service, the authors reviewed and edited the content as needed and take full responsibility for the content of the publication

\section*{Acknowledgments}

The funding support of the National Natural Science
Foundation of China (under the Grant Nos. 92052101 and
91952302) is acknowledged.

\appendix
\section{Hyperparameters in the Darcy problem}\label{ApDAR}
Here we list the hyperparameters of the networks with the best performance in the Darcy problem. Table \ref{tab:hyper-d-darcy}, \ref{tab:hyper-dd-darcy} and  \ref{tab:hyper-mdd-darcy} show the hyperparameters of DeepONet, Decoder-DeepONet and Multi-Decoder-DeepONet, respectively. 

\newcolumntype{Y}{>{\centering\arraybackslash}X}
\begin{table}[h]
        \caption{Hyperparameters of DeepONet}
	\centering
	\footnotesize
	\begin{tabularx}{\textwidth}{>{\hsize=.9\hsize\centering\arraybackslash}YYYYYY>{\hsize=.5\hsize\centering\arraybackslash}Y>{\hsize=.5\hsize\centering\arraybackslash}Y}
		\toprule
		
		\multirow{2}{*}{\textbf{subnet}} & \multirow{2}{*}{\textbf{Layer}} & \textbf{Input Shape} & \textbf{Output Shape} & \multirow{2}{*}{\textbf{Activation}} & \multirow{2}{*}{\textbf{Filters}} & \textbf{Filter Size} & \textbf{Stride Size}\\
		\midrule
		\multirow{5}{=}{Branch-net} & Conv2D & 61x61x1 & 29x29x64 & ReLU & 64 & 5x5 & 2 \\
		& Conv2D & 29x29x64 & 13x13x128 & ReLU & 128 & 5x5 & 2 \\
		& Flatten & 13x13x128 & 204928 & - & - & - & - \\
		& Dense & 204928 & 256 & ReLU & - & - & - \\
		& Dense & 256 & 256 & None & - & - & - \\
		\midrule
		\multirow{4}{=}{Trunk-net} & Dense & 2 & 256 & ReLU & - & - & - \\
		& Dense & 256 & 512 & ReLU & - & - & - \\
		& Dense & 512 & 512 & ReLU & - & - & - \\
		& Dense & 512 & 256 & None & - & - & - \\
		
		\bottomrule
	\end{tabularx}
	
	\label{tab:hyper-d-darcy}
\end{table}

\newcolumntype{Y}{>{\centering\arraybackslash}X}
\begin{table}[h]
        \caption{Hyperparameters of Decoder-DeepONet}
	\centering
	\footnotesize
	\begin{tabularx}{\textwidth}{>{\hsize=.9\hsize\centering\arraybackslash}YYYYYY>{\hsize=.5\hsize\centering\arraybackslash}Y>{\hsize=.5\hsize\centering\arraybackslash}Y}
		\toprule
		
		\multirow{2}{*}{\textbf{subnet}} & \multirow{2}{*}{\textbf{Layer}} & \textbf{Input Shape} & \textbf{Output Shape} & \multirow{2}{*}{\textbf{Activation}} & \multirow{2}{*}{\textbf{Filters}} & \textbf{Filter Size} & \textbf{Stride Size}\\
		\midrule
		\multirow{5}{=}{Branch-net} & Conv2D & 61x61x1 & 29x29x16 & ReLU & 16 & 5x5 & 2 \\
		& Conv2D & 29x29x16 & 13x13x8 & ReLU & 8 & 5x5 & 2 \\
		& Flatten & 13x13x8 & 128 & - & - & - & - \\
		& Dense & 128 & 200 & None & - & - & - \\
		\midrule
		\multirow{5}{=}{Trunk-net} & Conv2D & 61x61x1 & 29x29x16 & ReLU & 16 & 5x5 & 2 \\
		& Conv2D & 29x29x16 & 13x13x8 & ReLU & 8 & 5x5 & 2 \\
		& Flatten & 13x13x8 & 128 & - & - & - & - \\
		& Dense & 128 & 200 & None & - & - & - \\
		\midrule
		\multirow{3}{=}{Dot-net} & Flatten & 2x200 & 400 & - & - & - & - \\
		& Dense & 400 & 1000 & ReLU & - & - & - \\
		& Dense & 1000 & 3721 & None & - & - & - \\
		\bottomrule
	\end{tabularx}
	
	\label{tab:hyper-dd-darcy}
\end{table}

\newcolumntype{Y}{>{\centering\arraybackslash}X}
\begin{table}[h]
        \caption{Hyperparameters of Multi-Decoder-DeepONet}
	\centering
	\footnotesize
	\begin{tabularx}{\textwidth}{>{\hsize=.9\hsize\centering\arraybackslash}YYYYYY>{\hsize=.5\hsize\centering\arraybackslash}Y>{\hsize=.5\hsize\centering\arraybackslash}Y}
		\toprule
		
		\multirow{2}{*}{\textbf{subnet}} & \multirow{2}{*}{\textbf{Layer}} & \textbf{Input Shape} & \textbf{Output Shape} & \multirow{2}{*}{\textbf{Activation}} & \multirow{2}{*}{\textbf{Filters}} & \textbf{Filter Size} & \textbf{Stride Size}\\
		\midrule
		\multirow{5}{=}{Branch-net} & Conv2D & 61x61x1 & 29x29x16 & ReLU & 16 & 5x5 & 2 \\
		& Conv2D & 29x29x16 & 13x13x8 & ReLU & 8 & 5x5 & 2 \\
		& Flatten & 13x13x8 & 1352 & - & - & - & - \\
		& Dense & 1352 & 128 & ReLU & - & - & - \\
		& Dense & 128 & 20 & None & - & - & - \\
		\midrule
		\multirow{5}{=}{Branch-average-net} & Conv2D & 61x61x1 & 29x29x16 & ReLU & 16 & 5x5 & 2 \\
		& Conv2D & 29x29x16 & 13x13x8 & ReLU & 8 & 5x5 & 2 \\
		& Flatten & 13x13x8 & 1352 & - & - & - & - \\
		& Dense & 1352 & 128 & ReLU & - & - & - \\
		& Dense & 128 & 20 & None & - & - & - \\
		\midrule
		\multirow{5}{=}{Trunk-net} & Conv2D & 61x61x1 & 29x29x16 & ReLU & 16 & 5x5 & 2 \\
		& Conv2D & 29x29x16 & 13x13x8 & ReLU & 8 & 5x5 & 2 \\
		& Flatten & 13x13x8 & 1352 & - & - & - & - \\
		& Dense & 1352 & 128 & ReLU & - & - & - \\
		& Dense & 128 & 20 & None & - & - & - \\
		\midrule
		\multirow{3}{=}{Dot-net} & Flatten & 3x20 & 60 & - & - & - & - \\
		& Dense & 60 & 500 & ReLU & - & - & - \\
		& Dense & 500 & 3721 & None & - & - & - \\
		\bottomrule
	\end{tabularx}
	
	\label{tab:hyper-mdd-darcy}
\end{table}

\section{Validation of CFD Method} \label{ApA}
\begin{figure}[h]
	\centering
	\includegraphics[width=0.8\textwidth,trim=10 10 10 10,clip]{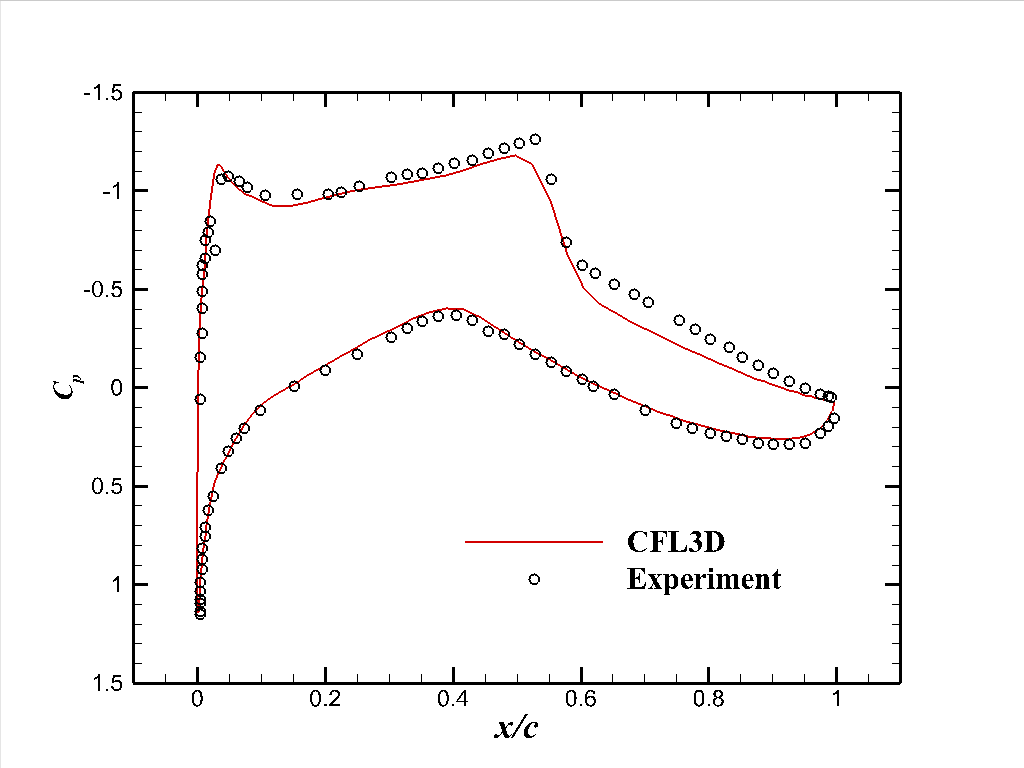}
	\caption{\label{fig:2822}The comparison of CFD results and experimental data }
\end{figure}

 We utilize the  experimental data of the RAE2822 Airfoil\cite{2822} to validate the accuracy of the CFD simulaiton employed in this study. The test conditions comprise a Mach number of 0.75, a Reynolds number of 6.5 million, and an angle of attack of 2.92 degrees. Figure \ref{fig:2822} shows the comparison of the CFD results and the experimental data. It can be seen that the simulated results have good agreements with the experimental data.

\section{Hyperparameters  in the prediction of flow-field around an airfoil}\label{ApAIR}
Here we list the hyperparameters of the networks with best performance, which are used in the prediction of flow-field around an airfoil. Talble \ref{tab:hyper-d-dg},  \ref{tab:hyper-d-sg},  \ref{tab:hyper-pod},  \ref{tab:hyper-dd}  and   \ref{tab:hyper-mdd} show the hyperparameters of DeepONet, DeepONet (aligned method), POD-DeepONet (aligned method), Decoder-DeepONet and Multi-Decoder-DeepONet, respectively.

\newcolumntype{Y}{>{\centering\arraybackslash}X}
\begin{table}[h]
        \caption{Hyperparameters of DeepONet}
	\centering
	\footnotesize
	\begin{tabularx}{\textwidth}{>{\hsize=.9\hsize\centering\arraybackslash}YYYYYY>{\hsize=.5\hsize\centering\arraybackslash}Y>{\hsize=.5\hsize\centering\arraybackslash}Y}
		\toprule
		
		\multirow{2}{*}{\textbf{subnet}} & \multirow{2}{*}{\textbf{Layer}} & \textbf{Input Shape} & \textbf{Output Shape} & \multirow{2}{*}{\textbf{Activation}} & \multirow{2}{*}{\textbf{Filters}} & \textbf{Filter Size} & \textbf{Stride Size}\\
		\midrule
		\multirow{5}{=}{Branch-net} & Conv2D & 32x16x1 & 14x6x64 & ReLU & 64 & 5x5 & 2 \\
		& Conv2D & 14x6x32 & 5x1x128 & ReLU & 128 & 5x5 & 2 \\
		& Flatten & 5x1x128 & 640 & - & - & - & - \\
		& Dense & 640 & 128 & ReLU & - & - & - \\
		& Dense & 128 & 200 & None & - & - & - \\
		\midrule
		\multirow{3}{=}{Trunk-net} & Dense & 2 & 256 & ReLU & - & - & - \\
		& Dense & 256 & 256 & ReLU & - & - & - \\
		& Dense & 256 & 200 & None & - & - & - \\
		
		\bottomrule
	\end{tabularx}
	
	\label{tab:hyper-d-dg}
\end{table}

\newcolumntype{Y}{>{\centering\arraybackslash}X}
\begin{table}[h]
        \caption{Hyperparameters of DeepONet (aligned method)}
	\centering
	\footnotesize
	\begin{tabularx}{\textwidth}{>{\hsize=.9\hsize\centering\arraybackslash}YYYYYY>{\hsize=.5\hsize\centering\arraybackslash}Y>{\hsize=.5\hsize\centering\arraybackslash}Y}
		\toprule
		
		\multirow{2}{*}{\textbf{subnet}} & \multirow{2}{*}{\textbf{Layer}} & \textbf{Input Shape} & \textbf{Output Shape} & \multirow{2}{*}{\textbf{Activation}} & \multirow{2}{*}{\textbf{Filters}} & \textbf{Filter Size} & \textbf{Stride Size}\\
		\midrule
		\multirow{5}{=}{Branch-net}  & Dense & 256 & 512 & ReLU & - & - & - \\
		& Dense & 512 & 1024 & ReLU & - & - & - \\
		& Dense & 1024 & 1024 & ReLU & - & - & - \\
		& Dense & 1024 & 512 & ReLU & - & - & - \\
		& Dense & 512 & 128 & None & - & - & - \\
		\midrule
		\multirow{3}{=}{Trunk-net} & Dense & 2 & 256 & ReLU & - & - & - \\
		& Dense & 256 & 256 & ReLU & - & - & - \\
		& Dense & 256 & 128 & None & - & - & - \\
		
		\bottomrule
	\end{tabularx}
	
	\label{tab:hyper-d-sg}
\end{table}

\newcolumntype{Y}{>{\centering\arraybackslash}X}
\begin{table}[h]
        \caption{Hyperparameters of POD-DeepONet (aligned method)}
	\centering
	\footnotesize
	\begin{tabularx}{\textwidth}{>{\hsize=.9\hsize\centering\arraybackslash}YYYYYY>{\hsize=.5\hsize\centering\arraybackslash}Y>{\hsize=.5\hsize\centering\arraybackslash}Y}
		\toprule
		
		\multirow{2}{*}{\textbf{subnet}} & \multirow{2}{*}{\textbf{Layer}} & \textbf{Input Shape} & \textbf{Output Shape} & \multirow{2}{*}{\textbf{Activation}} & \multirow{2}{*}{\textbf{Filters}} & \textbf{Filter Size} & \textbf{Stride Size}\\
		\midrule
		\multirow{5}{=}{Branch-net} & Conv2D & 32x16x1 & 14x6x64 & ReLU & 64 & 5x5 & 2 \\
		& Conv2D & 14x6x32 & 5x1x128 & ReLU & 128 & 5x5 & 2 \\
		& Flatten & 5x1x128 & 640 & - & - & - & - \\
		& Dense & 640 & 128 & ReLU & - & - & - \\
		& Dense & 128 & 128 & None & - & - & - \\
		\midrule
		\multirow{2}{=}{Trunk-net} & Dense & 2 & 128 & ReLU & - & - & - \\
		& Dense & 128 & 64 & None & - & - & - \\
		\midrule
	    \multicolumn{8}{c}{Number of POD modes: 64}\\
		\bottomrule
	\end{tabularx}
	
	\label{tab:hyper-pod}
\end{table}

\newcolumntype{Y}{>{\centering\arraybackslash}X}
\begin{table}[h]
        \caption{Hyperparameters of Decoder-DeepONet}
	\centering
	\footnotesize
	\begin{tabularx}{\textwidth}{>{\hsize=.9\hsize\centering\arraybackslash}YYYYYY>{\hsize=.5\hsize\centering\arraybackslash}Y>{\hsize=.5\hsize\centering\arraybackslash}Y}
		\toprule
		
		\multirow{2}{*}{\textbf{subnet}} & \multirow{2}{*}{\textbf{Layer}} & \textbf{Input Shape} & \textbf{Output Shape} & \multirow{2}{*}{\textbf{Activation}} & \multirow{2}{*}{\textbf{Filters}} & \textbf{Filter Size} & \textbf{Stride Size}\\
		\midrule
		\multirow{4}{=}{Branch-net} & Conv2D & 32x16x1 & 14x6x32 & ReLU & 32 & 5x5 & 2 \\
		& Conv2D & 14x6x32 & 5x1x16 & ReLU & 16 & 5x5 & 2 \\
		& Flatten & 5x1x16 & 80 & - & - & - & - \\
		& Dense & 80 & 200 & None & - & - & - \\
		\midrule
		\multirow{4}{=}{Trunk-net} & Conv2D & 100x241x2 & 33x79x32 & ReLU & 32 & 3x5 & 3 \\
		& Conv2D & 33x79x32 &  11x25x16 & ReLU & 16 & 3x5 & 3 \\
		& Flatten &  11x25x16 & 4400 & - & - & - & - \\
		& Dense & 4400 & 200 & None & - & - & - \\
		\midrule
		\multirow{5}{=}{Dot-net} & Conv2D-Transpose & \multirow{2}{*}{2x200x1} & \multirow{2}{*}{10x1000x64} & \multirow{2}{*}{ReLU} & \multirow{2}{*}{64} & \multirow{2}{*}{2x2} & \multirow{2}{*}{5} \\
		& Conv2D & 10x1000x64 & 10x1000x1 & ReLU & 1 & 1x1 & 1 \\
		& Flatten & 10x1000x1 & 10000 & - & - & - & - \\
		& Dropout & 10000 & 10000 & - & - & - & - \\
		& Dense & 10000 & 24100 & None & - & - & - \\
		\bottomrule
	\end{tabularx}
	
	\label{tab:hyper-dd}
\end{table}

\newcolumntype{Y}{>{\centering\arraybackslash}X}
\begin{table}[h]
        \caption{Hyperparameters of Multi-Decoder-DeepONet}
	\centering
	\footnotesize
	\begin{tabularx}{\textwidth}{>{\hsize=.9\hsize\centering\arraybackslash}YYYYYY>{\hsize=.5\hsize\centering\arraybackslash}Y>{\hsize=.5\hsize\centering\arraybackslash}Y}
		\toprule
		
		\multirow{2}{*}{\textbf{subnet}} & \multirow{2}{*}{\textbf{Layer}} & \textbf{Input Shape} & \textbf{Output Shape} & \multirow{2}{*}{\textbf{Activation}} & \multirow{2}{*}{\textbf{Filters}} & \textbf{Filter Size} & \textbf{Stride Size}\\
		\midrule
		\multirow{4}{=}{Branch-net} & Conv2D & 32x16x1 & 14x6x32 & ReLU & 32 & 5x5 & 2 \\
		 & Conv2D & 14x6x32 & 5x1x16 & ReLU & 16 & 5x5 & 2 \\
		 & Flatten & 5x1x16 & 80 & - & - & - & - \\
		 & Dense & 80 & 200 & None & - & - & - \\
		\midrule
		\multirow{4}{=}{Branch-average-net} & Conv2D & 100x241x1 & 48x119x32 & ReLU & 32 & 5x5 & 2 \\
		 & Conv2D & 48x119x32 & 22x58x16 & ReLU & 16 & 5x5 & 2 \\
		 & Flatten & 22x58x16 & 20416 & - & - & - & - \\
		 & Dense & 20416 & 200 & None & - & - & - \\
		\midrule
		\multirow{4}{=}{Trunk-net} & Conv2D & 100x241x2 & 48x119x32 & ReLU & 32 & 5x5 & 2 \\
		 & Conv2D & 48x119x32 &  22x58x16 & ReLU & 16 & 5x5 & 2 \\
		 & Flatten &  22x58x16 & 20416 & - & - & - & - \\
		 & Dense & 20416 & 200 & None & - & - & - \\
		\midrule
		\multirow{5}{=}{Dot-net} & Conv2D-Transpose & \multirow{2}{*}{3x200x1} & \multirow{2}{*}{15x1000x64} & \multirow{2}{*}{ReLU} & \multirow{2}{*}{64} & \multirow{2}{*}{2x2} & \multirow{2}{*}{5} \\
		 & Conv2D & 15x1000x64 & 15x1000x1 & ReLU & 1 & 1x1 & 1 \\
		 & Flatten & 15x1000x1 & 15000 & - & - & - & - \\
		 & Dropout & 15000 & 15000 & - & - & - & - \\
		 & Dense & 15000 & 24100 & None & - & - & - \\
		\bottomrule
	\end{tabularx}
	
	\label{tab:hyper-mdd}
\end{table}

\clearpage

\bibliographystyle{elsarticle-num-names} 
\bibliography{sample}





\end{document}